\documentclass{article} 

\usepackage[table]{xcolor}
\usepackage{iclr2024_conference,times}

\usepackage{amsmath,amsfonts,bm}

\def\eqref#1{equation~\ref{#1}}

\def\1{\bm{1}}

\DeclareMathAlphabet{\mathsfit}{\encodingdefault}{\sfdefault}{m}{sl}
\SetMathAlphabet{\mathsfit}{bold}{\encodingdefault}{\sfdefault}{bx}{n}

\newcommand{\E}{\mathbb{E}}

\usepackage{hyperref}
\usepackage{url}
\usepackage{graphicx}
\usepackage{cleveref}
\usepackage{booktabs}
\usepackage{multirow}
\usepackage{rotating}
\usepackage{enumitem}
\usepackage{floatrow}
\usepackage{blindtext}
\usepackage{makecell}

\newfloatcommand{capbtabbox}{table}[][\FBwidth]

\newcommand{\acr}[1]{{{\textsc{#1}}}}
\newcommand{\gqa}[0]{\texttt{GraphQA}}
\newcommand{\method}[1]{\acr{#1}}
\newcommand{\eg}[0]{\emph{e.g.},~}
\newcommand{\ie}[0]{\emph{i.e.},~}
\definecolor{lightgray}{gray}{0.95}
\renewcommand\fbox{\fcolorbox{cyan}{lightgray}}

\newcounter{insight}[section]
\newenvironment{myinsight}[1][\underline{Summary}]
{\vspace{-2pt}\refstepcounter{insight}\par\noindent\textbf{#1: }}{\par\vspace{-2pt}}

\title{Talk like a Graph: Encoding Graphs for \\Large Language Models}

\author{Bahare Fatemi, Jonathan Halcrow, Bryan Perozzi \\
Google Research\\
\texttt{\{baharef,halcrow,bperozzi\}@google.com} \\
}

\iclrfinalcopy 
\begin{document}

\maketitle

\begin{abstract}
Graphs are a powerful tool for representing and analyzing complex relationships in real-world applications such as social networks, recommender systems, and computational finance.
Reasoning on graphs is essential for drawing inferences about the relationships between entities in a complex system, and to identify hidden patterns and trends.
Despite the remarkable progress in automated reasoning with natural text, reasoning on graphs with large language models (LLMs) remains an understudied problem.
In this work, we perform the first comprehensive study of encoding graph-structured data as text for consumption by LLMs.
We show that LLM performance on graph reasoning tasks varies on three fundamental levels: (1) the graph encoding method, (2) the nature of the graph task itself, and (3) interestingly, the very structure of the graph considered.
These novel results provide valuable insight on strategies for encoding graphs as text. Using these insights we illustrate how the correct choice of encoders can boost performance on graph reasoning tasks inside LLMs by 4.8\% to 61.8\%, depending on the task.

\end{abstract}

\section{Introduction}

There has been remarkable recent progress in the research and applications of large language models~(LLMs)~\citep{vaswani2017attention, devlin2018bert, brown2020language, ouyang2022training}.
These generative models have captivated the artificial intelligence community and a plethora of models trained on a variety of tasks and modalities have recently been released \citep{zhao2023survey}.
All of these advancements have led to a growing consensus that LLMs are a pivotal advancement on the path to artificial general intelligence (AGI) \citep{bubeck2023sparks}.

However, despite all their successes, there are a number of limitations with the current methodology of design and implementation of LLMs.
One of the most obvious limitations is their reliance on unstructured text, causing the models to sometimes miss obvious logical entailments or \textit{hallucinate} incorrect conclusions \citep{zhang2023siren}.
Another is that LLMs are fundamentally limited by when they were trained, and it can be difficult to incorporate `fresh' information about the state of the world which has changed \citep{lewis2020retrieval}.
Graph-structured data is one of the most flexible ways to represent information and could be a promising solution to both challenges \citep{schneider2022decade, pan2023unifying}.

Interestingly, despite this promise, the intersection of graphs and LLMs has been relatively understudied.
For example, while much work has focused on LLMs and graph databases (or \textit{knowledge graphs} \citep{guu2020retrieval,lewis2020retrieval}) there has not been much study about general purpose use of graph-structured data.
More recently, \cite{wang2023can} have sought to address this by designing a graph benchmarking task for language models.
While their task represents an exciting initial foray into measuring LLMs graph reasoning capabilities, 
there are still many open questions due to the omission of several natural graph tasks and a lack of variety in the type of graph structure considered.
Other recent work seeks to replace graph-structured data with LLMs \citep{ye2023natural}, but this does not address fundamental challenges with LLMs.

In this work, we perform the first comprehensive study about reasoning over graph-structured data as text for consumption by LLMs.
To analyze graph reasoning more closely, we decompose the problem into \textit{graph encoding} and \textit{graph prompt engineering}.
Varying graph encoding methods allows us to understand how LLM's learned representations are leveraged in graph tasks.
While studying prompt engineering techniques finds the most suitable way to get a desired solution to a question from an LLM.
Our experimental results seek to uncover the situations where different prompt heuristics work well.
To that end, we propose a new set of benchmarks \gqa{} for measuring LLM performance reasoning over graph data.
\gqa{} is distinguished by using graphs with much more varied and realistic graph structure than has previously been studied with LLMs.

\textbf{Our Contributions}: Specifically, the contributions of our work are the following:

\begin{enumerate}[noitemsep,topsep=0pt]
    \item An \textbf{extensive study} of graph-structure prompting techniques for use in LLMs.
    \item \textbf{Insights and best practices} for encoding graphs as text for use in LLMs.
    \item A new \textbf{graph benchmark} (\gqa) to aid the community in studying the effects of graph structure on LLM prompting further.    
\end{enumerate}

\section{Prompting LLMs for Graph Reasoning}
\label{related_work}

\textbf{Notation.}
Let $f$ be the interface function to a generative AI model, which takes high-dimensional discrete input tokens $W$ and produces output in the same token space ($f: W \mapsto W$).
Without loss of generality, we will colloquially refer to $f$ as a pre-trained Large Language Model (LLM) throughout this work, but note that our discussion here applies to any generative AI model with such a discrete interface.
In this work, we consider encoding graphs $G = (V, E)$, where $V$ is the set of vertices (or nodes) and $E \in (V \times V)$ is the set of edges connecting them.

\begin{figure}[t]
  \vspace{-2em}
  \centering
  \includegraphics[width=0.8\textwidth]{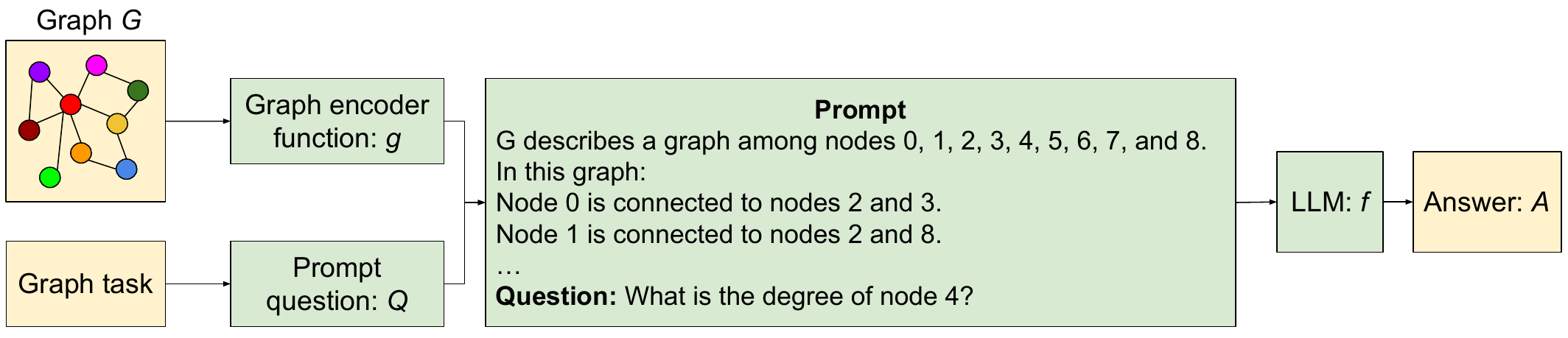}
  \caption{Overview of our framework for reasoning with graphs using LLMs.
    }\label{fig:main_framework}
\end{figure}

\subsection{Prompt Engineering}

The goal in prompt engineering is to find the correct way to phrase a question $Q$ such that an LLM $f$ (or other generative model) will return the corresponding answer $A$, $(Q \in W, A \in W$).  In other words:
$$A = f(Q)$$
In this work, our goal is to provide the LLM $f$ with graph information, so that it can better reason about question/answer pairs that require access to arbitrarily structured relational information.
$$A = f(G, Q)$$
A variety of approaches exist for modifying the LLM $f(.)$ so that it could better perform on tasks with graph data such as fine-tuning \citep{clark2020transformers}, soft prompting \citep{lester2021power}, and LoRA \citep{hu2021lora}.
In addition, many approaches modify the model to include graph information~\citep{muller2023attending,zhang2020graph,dwivedi2020generalization}.
However, these methods all require access to the internals of the model (either its weights or gradients), which can limit their applicability in many real-world settings.
In this work, we are instead interested in the case where $f(.)$ and its parameters are fixed, and the system is available only for use in a \textit{black box} setup where the LLM only consumes and produces text (\ie the LLM $f: W \mapsto W$).
We believe this setting to be particularly valuable as the number of proprietary models available and their hardware demands increase.

To this end, we introduce the graph encoding function $g(G)$ and question rephrasing function $q(Q)$, where  $g: G \mapsto W$ and $q: W \mapsto W$ (where $W$ is the large discrete domain of tokens used to train the LLM). 
\begin{equation}
    A = f(g(G), q(Q))
\end{equation}
Our training input $D$ to the graph-based prompt system is a set of ${G, Q, S}$ triples, where $G$ is a graph, $Q$ is a question, and $S$, $S \in W$, is a solution to $Q$.  We seek to find a $g(.)$ and $q(.)$ that maximize the expected score from the model ($\text{score}_f$) of the answers over the training dataset $D$.
\begin{equation}
  \label{eq:prompting}
  \max_{g, q} \E_{G, Q, S \in D} \text{ score}_f(g(G), q(Q), S)  
\end{equation}
As $W$ is a very large discrete space, many current approaches use heuristics for this optimization~(by changing the prompt $Q$).
The novel contribution of this work is to consider the role of the graph encoding function $g(.)$, $q(.)$ question rephrasing function, and the graph structure $G$ in the optimization of \Cref{eq:prompting}.

\subsection{Prompting Heuristics}

The vast majority of prompting heuristics operate by optimizing the prompt text $Q$ used to query the model.
We briefly introduce the methods we examine further in the paper here:\\
\textbf{Zero-shot prompting}~({\method{zero-shot}}): This approach simply provides the model with a task description and asks it to generate the desired output, without any prior training on the task.
\textbf{Few-shot in-context learning}~({\method{few-shot}})~\citep{fewshot}: This approach provides the model with a small number of examples of the task, along with the desired outputs. The model then learns from these examples to perform the task on new inputs.
\textbf{Chain-of-thought}~(CoT) prompting ({\method{cot}})~\citep{wei2022chain}: This approach provides the model with a sequence of examples, each of which shows how to solve the task step-by-step. The model then learns to generate its CoTs to solve new problems. 
\textbf{Zero-shot CoT} prompting   ({\method{zero-cot}})~\citep{kojima2022large}: This approach is similar to CoT prompting, but it does not require any prior training examples. Instead, the model uses a simple prompt to generate its own CoTs. As suggested by the original paper, we used ``Let's think step by step''.
\textbf{Bag prompting} ({\method{cot-bag}})~\citep{wang2023can}: This technique is proposed to improve the performance of LLMs on graph-related tasks. It works by appending ``Let's construct a graph with the nodes and edges first'' to the graph description.

We note that there is also a popular recent extension of this family of methods, based on \textit{iterative prompting}. These methods use a series of iterative LLM queries to optimize the prompt question~(\eg \citep{zhou2022large,pryzant2023automatic,yang2023large}).
However, our initial experiments showed that iterative prompting methods performed much worse for our tasks, due to cascading errors. Consequently, we chose to concentrate our efforts on the methods outlined above.


In this study, the goal is to optimize the graph encoding function on basic graph tasks. Such basic tasks are essential intermediate steps for more complex reasoning tasks on graphs. 
We conduct extensive experiments on graph encoding function, question, and graph generator functions, providing a study of graph encoding methods for black-box LLM usage.


\section{Talk Like a Graph: Encoding Graphs via Text}

Graph encoding is a necessary step for turning graph-structured information into a sequence for consumption by language models.
In this section, we will study the details of a graph encoding function $g(.)$ which maps graph data into tokens for consumption by an LLM.
Our experimental results in this section seek to understand the best form of graph encoding and prompt engineering to maximize the performance on graph reasoning tasks.

We begin by highlighting some of the most exciting results from our analysis here:
\begin{itemize}[noitemsep,topsep=0pt]
    \item \textbf{R1}: LLMs perform poorly on basic graph tasks (\S \ref{sec:graph-encoders}).
    \item \textbf{R2}: The graph encoding function has a significant impact on LLM graph reasoning (\S \ref{sec:graph-encoders}).
    \item \textbf{R3}: Model capacity has a significant effect on graph reasoning capabilities of LLMs (\S \ref{sec:model-capacity}).
\end{itemize}

\begin{figure}[t]
  \vspace{-2em}
  \centering
  \includegraphics[width=0.85\textwidth]{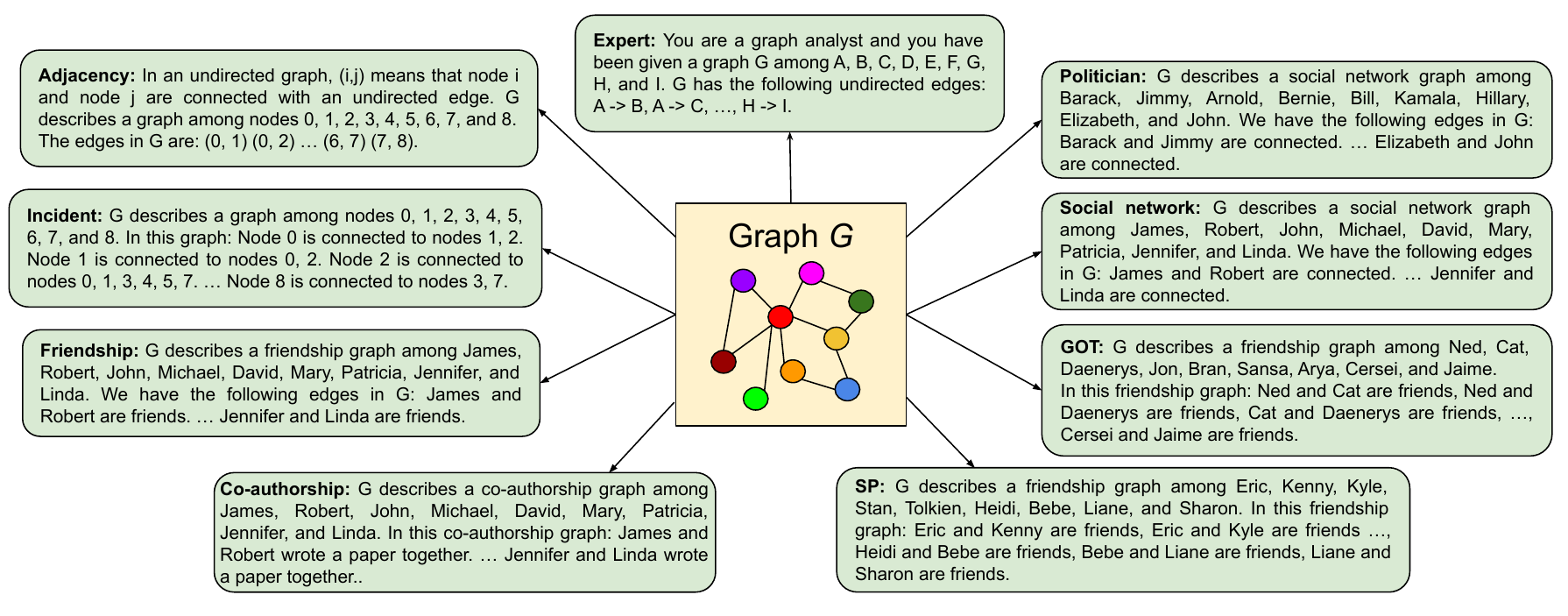}
  \caption{Overview of our framework for encoding graphs via text.
    }\label{fig:graph_encoder_function}
\end{figure}

\textbf{Graph encoding function.}
This section is an investigation into various methodologies for representing graphs as text. 
This process of encoding graphs as text can be separated into two key inquiries:
First, the encoding of nodes in the graph, and second the encoding of edges between the nodes. Regarding the encoding of nodes and edges, we examine several techniques. \Cref{fig:graph_encoder_function} shows an overview of the graph encoding functions used.
For brevity's sake, a full description and examples of the graph encoding functions considered are explained in \Cref{sec:graph-encoding-detail}.

\textbf{Graph structure.}
We briefly note that the design of this experiment follows that of~\cite{wang2023can}, who use Erd\H{o}s-R\'enyi~(ER) graphs~\citep{erdds1959random}.
One contribution of our work is to consider the effect of more complex graph structures on reasoning in LLMs (covered in \Cref{sec:graph_generation}).

\subsection{Experiment 1: Varying graph encoding functions}\label{sec:graph-encoders}
In this experiment, we measure the performance of pre-trained LLMs on graph tasks: \textit{edge existence}, \textit{node degree}, \textit{node count}, \textit{edge count}, \textit{connected nodes}, and \textit{cycle check}. We describe these tasks and our graph benchmark that contains them (\gqa) in detail in \Cref{sec:basic-graph-tasks}.

\subsubsection{Results}

\Cref{table:graph-encoders-palm62} shows the results of this experiment varying graph encoding and prompting techniques.
These results show several interesting conclusions, which we briefly summarize here:

\begin{table}[t]
\vspace{-2em}
\centering
\footnotesize
\setlength{\tabcolsep}{2pt}
\resizebox{\columnwidth}{!}{%
\setlength{\tabcolsep}{3pt}
\begin{tabular}{c|c|cccccc}
\textbf{Method} & \textbf{Encoding} & \textbf{Edge Existence} & \textbf{Node degree} & \textbf{Node count} & \textbf{Edge count} & \textbf{Connected nodes} & \textbf{Cycle check} \\ \hline
\multirow{9}{*}{\begin{sideways}\method{\normalsize zero-shot}\end{sideways}}
&  \cellcolor{gray!25} Overall ($\mu / \delta$) &\cellcolor{gray!25} \underline{44.5} / 9.4 &\cellcolor{gray!25} 14.0/16.0 &\cellcolor{gray!25} 21.73 / 8.6 &\cellcolor{gray!25} 12.4 / 4.8 &\cellcolor{gray!25} 14.7 / 11.0 &\cellcolor{gray!25} \underline{76.0} / 13.2\\
& Adjacency & 45.8 & 12.4 & 18.8 & 14.0 & 19.8 & 71.6\\
& Incident & 39.6 & 25.0 & 15.6 & 10.6 & 53.8 & 68.8\\
& Co-authorship & 44.0 & 13.8 & 22.0 & 11.4 & 7.6 & 70.8\\
& Friendship & 46.6 & 11.2 & 23.0 & 10.2 & 4.0 & \textbf{82.0}\\
& SP & 46.4 & 9.0 & 22.4 & 15.0 & 6.2 & 80.4\\
& GOT & \textbf{49.0} & 13.6 & 22.8 & 13.2 & 7.6 & 79.0\\
& Social network & 43.2 & 16.0 & 22.8 & 10.8 & 8.2 & 81.2\\
& Politician & 44.6 & 15.2 & 24.2 & 11.6 & 8.8 & 81.0\\
& Expert & 41.2 & 10.0 & 24.0 & 14.8 & 16.4 & 69.6\\
\cline{2-8}
\multirow{9}{*}{\begin{sideways}\method{\normalsize zero-cot}\end{sideways}} 
& \cellcolor{gray!25} Overall ($\mu / \delta$) &\cellcolor{gray!25} 33.5 / 11.6 &\cellcolor{gray!25} 10.4 / 22.4 &\cellcolor{gray!25} 14.6 / 9.4 &\cellcolor{gray!25} 9.4 / 4.8 &\cellcolor{gray!25} 8.8 / 9.2 &\cellcolor{gray!25} 32.3 / 23.2\\
& Adjacency & 34.2 & 15.4 & 11.0 & 12.2 & 6.0 & 46.2\\
& Incident & 41.4 & 26.6 & 10.0 & 12.2 & 35.2 & 39.0\\
& Co-authorship & 29.8 & 9.8 & 15.6 & 8.2 & 3.0 & 28.2\\
& Friendship & 28.4 & 7.0 & 19.4 & 7.4 & 3.0 & 31.2\\
& SP & 32.6 & 9.2 & 15.6 & 8.4 & 5.0 & 34.8\\
& GOT & 34.6 & 8.4 & 16.2 & 8.4 & 5.4 & 33.4\\
& Social network & 30.8 & 6.6 & 14.0 & 9.2 & 3.8 & 26.0\\
& Politician & 38.0 & 4.2 & 14.6 & 8.6 & 3.2 & 23.0\\
& Expert & 31.6 & 6.0 & 14.8 & 10.0 & 14.2 & 28.8\\
\cline{2-8}
\multirow{9}{*}{\begin{sideways}\method{\normalsize few-shot}\end{sideways}} 
& \cellcolor{gray!25} Overall ($\mu / \delta$) &\cellcolor{gray!25} 36.8 / 13.8 &\cellcolor{gray!25} 17.4 / 23.4 &\cellcolor{gray!25} 25.3 / 35.6 &\cellcolor{gray!25} 12.0 / 9.0 &\cellcolor{gray!25} 12.4 / 15.2 &\cellcolor{gray!25} 37.4 / 24.0\\
& Adjacency & 42.8 & 15.4 & 47.2 & 18.6 & 22.2 & 47.8\\
& Incident & 38.8 & 33.6 & 51.2 & 14.6 & 36.6 & 45.0\\
& Co-authorship & 29.4 & 15.6 & 15.6 & 10.2 & 9.0 & 46.8\\
& Friendship & 40.6 & 12.2 & 18.4 & 9.8 & 6.4 & 41.4\\
& SP & 34.6 & 18.0 & 18.0 & 12.0 & 6.8 & 38.2\\
& GOT & 40.6 & 17.2 & 14.2 & 12.0 & 3.4 & 28.6\\
& Social network & 37.4 & 15.0 & 21.2 & 10.2 & 7.8 & 34.2\\
& Politician & 38.0 & 13.4 & 21.4 & 9.6 & 7.8 & 30.8\\
& Expert & 29.0 & 16.6 & 20.4 & 11.2 & 11.8 & 23.8\\
\cline{2-8}
\multirow{9}{*}{\begin{sideways}\method{\normalsize cot}\end{sideways}}
& \cellcolor{gray!25} Overall ($\mu / \delta$)&\cellcolor{gray!25} 42.8 / 7.0 &\cellcolor{gray!25} \underline{29.2} / 60.4 &\cellcolor{gray!25} \underline{27.6} / 42.4 &\cellcolor{gray!25} \underline{12.8} / 17.4 &\cellcolor{gray!25} 13.1 / 18.0 &\cellcolor{gray!25} 58.0 / 16.4 \\
& Adjacency & 42.8 & 71.2 & 57.0 & \textbf{25.2} & 22.4 & 56.6\\
& Incident & 41.6 & \textbf{75.0} & \textbf{57.6} & 21.4 & 30.2 & 62.6\\
& Co-authorship & 43.2 & 16.4 & 15.2 & 8.8 & 8.4 & 54.8\\
& Friendship & 46.6 & 14.6 & 23.0 & 7.8 & 9.6 & 61.8\\
& SP & 42.6 & 17.4 & 17.0 & 10.6 & 8.2 & 59.4\\
& GOT & 44.0 & 17.8 & 16.2 & 11.8 & 7.2 & 60.4\\
& Social network & 42.6 & 16.4 & 21.6 & 8.4 & 8.0 & 60.6\\
& Politician & 42.2 & 16.6 & 22.6 & 9.2 & 9.4 & 59.4\\
& Expert & 39.6 & 17.4 & 18.0 & 12.4 & 14.4 & 46.2\\
\cline{2-8}
\multirow{9}{*}{\begin{sideways}\method{\normalsize cot-bag}\end{sideways}} 
& \cellcolor{gray!25} Overall ($\mu / \delta$)&\cellcolor{gray!25} 37.3 / 16.6 &\cellcolor{gray!25} 28.0 / 61.8 &\cellcolor{gray!25} 26.9 / 33.8 &\cellcolor{gray!25} 12.5 / 17.8 &\cellcolor{gray!25} \underline{15.8} / 31.8 &\cellcolor{gray!25} 52.1 / 26.0 \\
& Adjacency & 45.8 & 66.8 & 48.6 & 25.0 & 20.6 & 56.8\\
& Incident & 45.6 & 75.2 & 51.2 & 21.8 & \textbf{41.0} & 63.0\\
& Co-authorship & 25.0 & 14.6 & 17.4 & 7.2 & 9.2 & 37.0\\
& Friendship & 39.0 & 16.2 & 21.8 & 7.4 & 9.8 & 52.0\\
& SP & 33.6 & 17.0 & 21.6 & 11.4 & 11.4 & 52.2\\
& GOT & 32.6 & 15.6 & 18.0 & 11.0 & 10.0 & 54.6\\
& Social network & 44.8 & 13.4 & 19.6 & 9.0 & 10.0 & 51.2\\
& Politician & 40.4 & 17.6 & 22.8 & 8.2 & 10.2 & 57.2\\
& Expert & 29.2 & 15.8 & 20.8 & 11.6 & 20.4 & 45.0\\
\end{tabular}
}
\caption{Comparison of various graph encoder functions based on their accuracy on different graph tasks using PaLM 62B. The most effective prompting heuristic is highlighted with an underline, and the top-performing graph encoder function for it is highlighted in bold. The overall result is represented its average~($\mu$) and an absolute difference~($\delta$) of its best and worst graph encoder.}
\label{table:graph-encoders-palm62}
\end{table}

\textbf{LLMs perform poorly on basic graph tasks.} Let's start by examining the overall results.
LLMs performed poorly on almost all the basic graph tasks we experimented with. 
This is especially interesting for the edge existence and cycle check tasks, where there is not an edge $53.96\%$ of the time for the edge existence task and there is a cycle $81.96\%$ of the time for the cycle check task. Therefore. LLMs perform worse than the majority baseline. Note that we experimented with ER graphs in this experiment, and it is very likely for an ER graph to have a cycle.

\textbf{Simple prompts are best for simple tasks.}
We see that \method{zero-cot} prompting has worse model performance than \method{zero-shot} prompting on basic graph tasks. This is likely because \method{zero-shot} prompting is sufficient for these tasks, which do not require multi-hop reasoning. \method{zero-cot} prompting can be effective for tasks that require multi-hop reasoning, such as arithmetic problems, but it is not necessary for most basic graph tasks, which only require the LLM to have an understanding of the graph structure (nodes, edges, paths, etc.) and the graph task.
However for more complex tasks,
adding few-shot examples and CoT prompting generally improved the performance of the model. This is mainly because few-shot examples provide the LLM with a better understanding of the task it is solving. CoT prompting can also improve performance by helping the LLM to find out how to get to the answer to the problem.

\textbf{graph encoding functions have significant impact on LLM reasoning.}
As the results indicate, the choice of the graph encoding function has a significant impact on the performance of LLMs on graph-related tasks. This is because different encoder functions capture different aspects of the graph structure.
For instance, for finding connected nodes to a node in a graph, adjacency achieves $19.8\%$ accuracy and incident achieves $53.8\%$ accuracy. 
For both node degree and connected nodes, incident encoding outperforms the rest of the encoder functions. This is likely because the incident encoder encodes the graph structure in a way that makes the relevant information more accessible, \ie in close proximity, to the LLM.

\textbf{Integer node encoding improves arithmetic performance.}
Another finding here is that integer encoding of nodes (\eg \emph{node 0}) can improve the performance of LLMs on integer output tasks, such as predicting node degree, node count, and edge count. This is because the input and output of the LLM are then in the same space, making it easier for the model to learn the relationship between the two.
Interestingly however, encoder functions with specific names (\eg David) worked better in non-integer output tasks such as \texttt{GOT} for edge existence or \texttt{Friendship} for cycle check.

\begin{myinsight}
Choosing the right graph encoding function significantly affects the performance of LLMs on basic graph algorithms. Therefore, it is important to select a function carefully and appropriately for the specific task. This finding is especially important because many reasoning tasks involve graph problems. For example, finding influential nodes in a social network is similar to finding the degree of the nodes in the graph. Encoding such graphs in the right way for the task can improve the task.  We examine the relative rankings of graph encodings more in \Cref{sec:encoding_ranking}.
\end{myinsight}

\subsection{Experiment 2: Varying Prompt Questions}
In this experiment, we maintained the graph encoding function as a constant for the concept of friendship and conducted experiments using two distinct question encoder functions: the graph question encoder and the application question encoder. The graph question encoder is responsible for encoding graph-related tasks, such as determining the degree of a specific node (\eg ``What is the degree of node $i$?''). This encoder is used for obtaining results in \Cref{sec:graph-encoders}.
On the other hand, the application question encoder interprets graph questions in a more practical, day-to-day context. In the application scenario, we used a friendship-based scenario where we transformed the tasks as follows: 
\emph{edge existence} became ``assessing friendship existence'', \emph{node degree} became ``counting the number of friends for an individual'', \emph{node count} became ``counting the number of people mentioned'', \emph{edge count} became ``counting the number of friendships mentioned'', and \emph{connected nodes} became ``listing friends''.

\textbf{Results:}
\Cref{table:graph-application} summarizes the results of our experiment on question encoder functions. As the results show, the application encoder outperforms the graph encoding on almost all tasks, despite both encoders having the same graph encoding function and only differing slightly in how they ask the question. For example, on the \method{zero-shot} edge existence task using PALM 2 XXS, the graph encoding obtained $42.8\%$ accuracy, while the application encoder obtained $60.8\%$.

\begin{myinsight}
The selection of the question encoder function affects the performance of LLMs when handling basic graph algorithms. As a result, it becomes important to translate a given task into more contextually meaningful textual information when employing LLMs for inference.
\end{myinsight}


\begin{table}[t!]
\centering
\footnotesize
\setlength{\tabcolsep}{2pt}
\resizebox{\columnwidth}{!}{%
\setlength{\tabcolsep}{3pt}
\begin{tabular}{cccccccc}
\textbf{Method} & \textbf{Question encoder} & \textbf{LLM} & \textbf{Edge Existence} & \textbf{Node degree} & \textbf{Node count} & \textbf{Edge count} & \textbf{Connected nodes }\\ \hline
\multirow{4}{*}{\method{zero-shot}}
& \cellcolor{gray!25}Graph &\cellcolor{gray!25} PaLM 2-XXS &\cellcolor{gray!25} 42.8 &\cellcolor{gray!25} 10.8 &\cellcolor{gray!25} 5.4 &\cellcolor{gray!25} \textbf{5.6} &\cellcolor{gray!25} 1.6 \\
& Application & PaLM 2-XXS & \textbf{60.8} & \textbf{14.0} & \textbf{9.4} & 4.4 & \textbf{11.4} \\ 
& \cellcolor{gray!25}Graph &\cellcolor{gray!25} PaLM 62B &\cellcolor{gray!25} 46.6 &\cellcolor{gray!25} 11.2 &\cellcolor{gray!25} \textbf{23.0} &\cellcolor{gray!25} 10.2 &\cellcolor{gray!25} 4.0 \\
& Application & PaLM 62B & \textbf{47.8} & \textbf{16.6} & 17.8 & \textbf{13.2} & \textbf{6.0} \\
\hline
\multirow{4}{*}{\method{cot}}
& \cellcolor{gray!25}Graph &\cellcolor{gray!25} PaLM2 XXS &\cellcolor{gray!25} 50.4 &\cellcolor{gray!25} 8.8 &\cellcolor{gray!25} 8.4 &\cellcolor{gray!25} 4.2 &\cellcolor{gray!25} 10.2 \\
& Application & PaLM2 XXS & \textbf{56.4} & \textbf{12.2} & \textbf{8.6} & \textbf{5.4} & \textbf{11.0} \\ 
& \cellcolor{gray!25}Graph &\cellcolor{gray!25} PaLM 62B &\cellcolor{gray!25} \textbf{46.6} &\cellcolor{gray!25} 14.6 &\cellcolor{gray!25} \textbf{23.0} &\cellcolor{gray!25} 7.8 &\cellcolor{gray!25} 9.6 \\
& Application & PaLM 62B & 38.6 & \textbf{16.6} & 16.0 & \textbf{12.2} & \textbf{10.0}  \\
\end{tabular}
}
\caption{Comparing two question encoders based on their accuracy for PaLM 2 XXS and PaLM 62B. The top-performing question encoder for the respective LLM is highlighted in bold.}
\label{table:graph-application}
\end{table}

\subsection{Experiment 3: Multiple Relation Encoding}
In this experimental setup, we introduce a modification to the friendship graph encoding function, which characterizes edges based on a range of distinct relation types, including  
\emph{friends}, \emph{colleagues}, \emph{spouses}, \emph{siblings}, \emph{neighbors}, \emph{acquaintances}, \emph{teammates}, \emph{classmates}, \emph{coworkers}, or \emph{roommates}. The selection of the relation type is randomized from this predefined set, thereby using multiple words to reference the existence of a relationship between nodes. 
This is a departure from using the same token(s) for edge representation in prior graph encoding experiments.

\begin{figure}
\begin{floatrow}

\capbtabbox{%
\resizebox{0.4\textwidth}{!}{%

\begin{tabular}{cccc}
 &\textbf{Task} & \makecell{\textbf{Same} \\ \textbf{relation}} & \makecell{\textbf{Multiple} \\ \textbf{relations}} \\ \hline
 \multirow{5}{*}[-0.5em]{\begin{sideways}\method{zero-shot}\end{sideways}}
&Edge Existence & \textbf{42.8} & 39.8 \\
&Node degree & 10.8 & \textbf{11.6}  \\
&Node count & 5.4 & \textbf{6.6}  \\
&Edge count & \textbf{5.6} & 5.4  \\
&Connected nodes & 1.6 & \textbf{3.4}  \\
&Cycle Check &  65.2 & \textbf{84.4} \\

\hline
 \multirow{6}{*}{\begin{turn}{90}\method{cot}\end{turn}}

&Edge Existence & 50.4 & \textbf{50.8} \\
&Node degree &  8.8 & \textbf{10.0} \\
&Node count & \textbf{8.4} & 5.8 \\
&Edge count  & 4.2 & \textbf{5.0} \\
&Connected nodes  & \textbf{10.2} & 7.2 \\
&Cycle Check & \textbf{77.4} & 74.4 \\
\\
\\
\\
\end{tabular}

}}{%
\caption{Results on multiple relations for edge encoding with PaLM 2 XXS. 
}
\label{table:random-relation}
}

\ffigbox{%
  \includegraphics[width=0.5\textwidth]{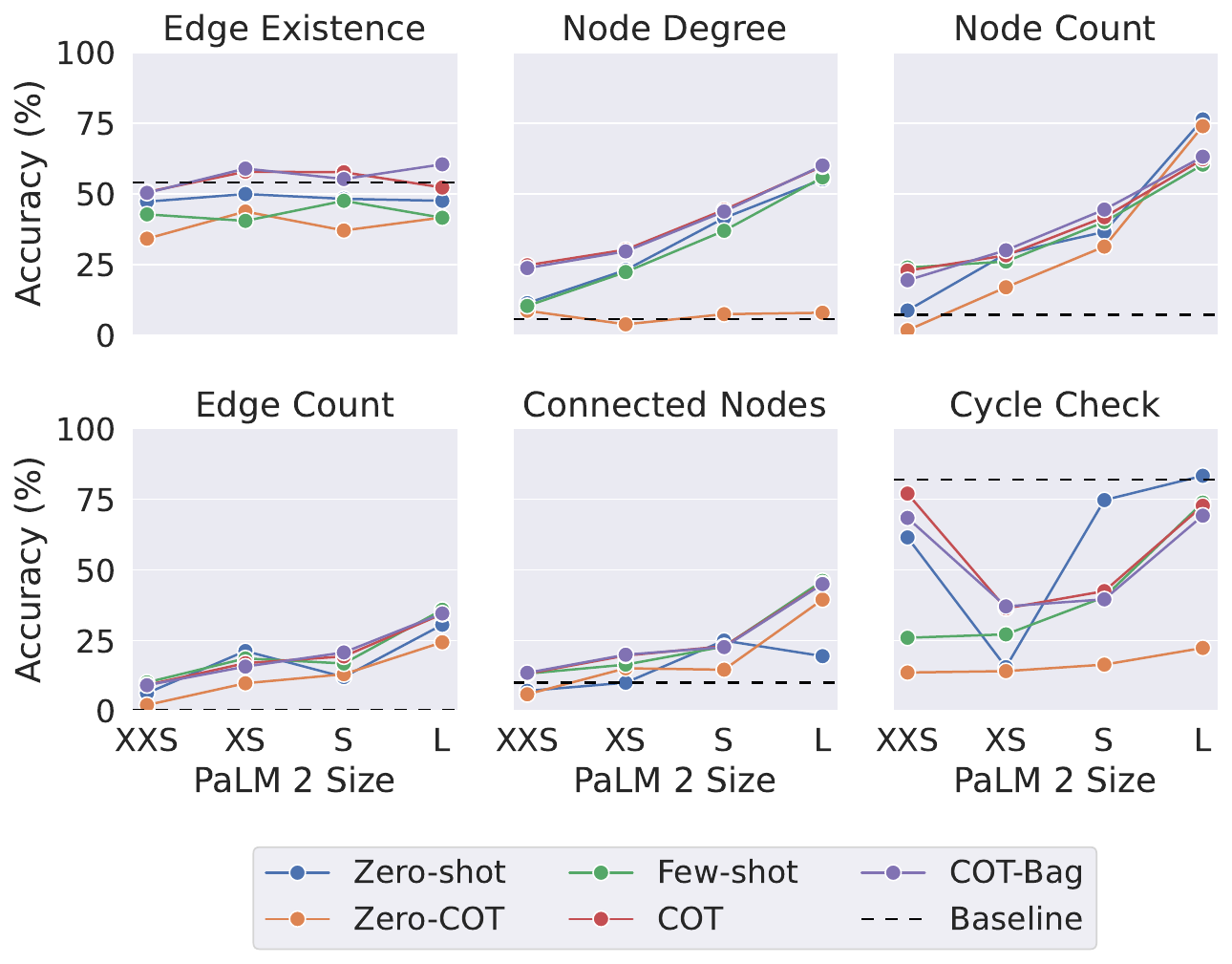}
}{%
  \caption{Effect of Model Capacity on graph reasoning task for PaLM 2-XXS, XS, S, and L.}
  \label{fig:model-capacity}
}

\end{floatrow}
\end{figure}

\textbf{Results:}
As Table~\ref{table:random-relation} shows, using multiple words to represent relationships did not hurt LLM performance and even improved it in some cases. This improvement is likely because the diverse set of relations provides the LLM with more textual information to perform the task, and the final encoding is closer to the text that the LLM may have seen during training, compared to the prior setup.

\subsection{Experiment 4: Model Capacity and Graph Reasoning Ability}\label{sec:model-capacity}
In this experiment, we measure the effect of model capacity on the graph tasks. We compare the results of PaLM~2~\citep{anil2023palm} XXS, XS, S, and L, which have different number of parameters and therefore different capacity. We report the majority baseline for reference.

\textbf{Results: Model capacity has a significant effect on the graph reasoning ability of an LLM.}
The results of this experiment, reported in \Cref{fig:model-capacity}, show the larger model is generally better at graph reasoning tasks. This is because it has more capacity to learn and store complex information. The model capacity has less effect on edge existence. The results also show that the model was not able to beat the majority baseline for edge existence even with a large capacity.

\begin{figure}[t]
  
\end{figure}

\subsection{Experiment 5: Reasoning in the Absence of Edges}
In this experiment, we evaluate the performance of LLMs on the \emph{disconnected nodes} task. This task differs from the previous ones in that it requires reasoning about information that is implicit in the graph, \ie information that is not explicitly mentioned in the output of the graph encoding function.

\textbf{Results: LLMs lack a global model of a graph.}
The \method{zero-shot} prompting method achieved an accuracy of $0.5\%$, while the \method{zero-cot}, \method{few-shot}, \method{cot}, and \method{cot-bag} methods achieved close to $0.0\%$ accuracy. These results suggest that LLMs perform significantly worse on the disconnected nodes task than on the connected nodes task.
We believe that this is because the graph encoding functions primarily encode information about connected nodes, while not explicitly encoding information about nodes that are not connected. As a result, LLMs are better at processing relationships among connected nodes than at capturing the absence of connections, leading to sub-optimal performance in disconnectivity-related tasks.

\section{Does the structure of the graph matter for the LLM?}\label{sec:graph_generation}

\begin{figure}[t]
  \vspace{-2em}
  \centering
  \includegraphics[width=1.0\textwidth]{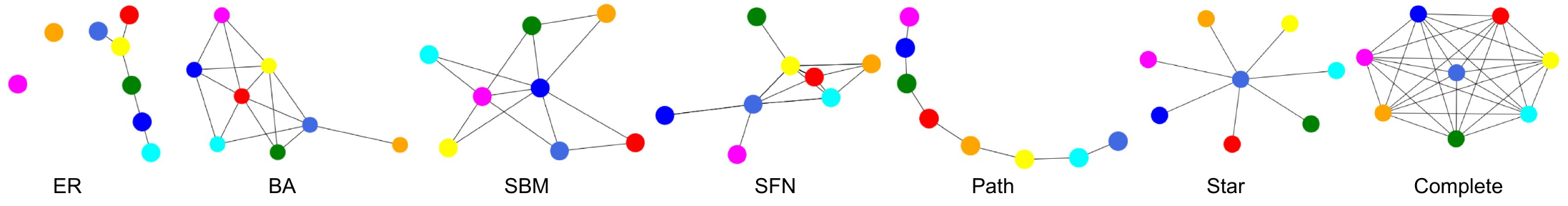}
  \caption{Samples of graphs generated with different graph generators in our framework.
    }\label{fig:graph_generators}
\end{figure}

It is natural to wonder if the structure of the graph itself might effect LLM's ability to reason over it. 
Inspired by recent work in analyzing graph neural networks \citep{palowitch2022graphworld,yasir2023examining} this section seeks to measure a LLM's reasoning capabilities over graph with distinct structures.
In this section, we show that graph structure can have significance influence on an LLM's reasoning performance. 
\Cref{fig:graph_generators} illustrates graphs created through different generative processes.

\subsection{Random Graph Generation}
To be able to experiment with LLMs on graphs, we generate random graphs using various graph generator algorithms. This allows us to:

\textbf{Cover a wide range of properties.} Different graph generators produce graphs with different properties. For example, Erdős-Rényi graphs tend to be sparse and have a small average degree, while Barabási-Albert graphs tend to be dense and have a power-law degree distribution. By using a diverse set of generators, we ensure that the \gqa{} benchmark includes graphs with a wide range of properties.

\textbf{Avoid bias in graph problem evaluation.} The goal of generating such graphs is to test the ability of LLMs to solve graph problems. Graph problems can vary in difficulty depending on the properties of the graphs, so we use a diverse set of graphs to avoid bias.

\textbf{Provide realistic benchmarks.} Real-world graphs exhibit a wide range of properties, and no single graph generator can capture all of these properties perfectly. By using a diverse set of generators, we create a benchmark that is more representative of real-world graphs.

To generate random graphs, we use Erd\H{o}s-R\'enyi~(ER) graphs~\citep{erdds1959random}, scale-free networks~(SFN)~\citep{scalefree}, Barabási–Albert~(BA) model~\citep{barabasi}, and stochastic block model~(SBM)~\citep{holland1983stochastic}, in addition to star, path, and complete graph generators. We use NetworkX~\citep{networkx} to generate the random graphs. The details are reported in \Cref{sec:implementation-detail}.

\subsection{Results on Random Graph Generators}

Previous experiments have studied the performance of LLMs on basic graph tasks using random graphs generated using the Erd\H{o}s-R\'enyi~(ER) model. 
However, ER graphs often do not accurately represent the characteristics of real-world graphs. In this experiment, we investigate the effect of different random graph generators on the performance of LLMs on graph reasoning tasks. To make the experiment more realistic, we sample the few-shot examples randomly from graphs generated using different algorithms. We report the results of this experiment in Table~\ref{table:graph-generators}. 

\textbf{Graph structure has a significant impact on the LLM's performance.} The results show that the algorithm used to generate the graph has a significant impact on the performance of the LLM on graph tasks. 
For example, the cycle check task achieves $91.7\%$ accuracy on complete graphs and $5.9\%$ accuracy on path graphs. This is because the LLM has a strong prior towards graphs having cycles. Therefore, the accuracy is high for complete graphs, which always have cycles, and very low for path graphs, which never have cycles. By adding few-shot examples some having a cycle and some not, the accuracy of cycle check on path graphs increased from $5.9\%$ to $19.7\%$. 
As another example, on the edge existence task, the LLM achieves $60.0\%$ accuracy on path graphs, which are less likely to have an edge between two nodes, and $19.8\%$ accuracy on complete graphs, which have edges between all pairs of nodes. This shows that the LLM has a prior that two nodes in a graph are more likely to be disconnected.

\begin{table}[t]
\centering
\footnotesize
\setlength{\tabcolsep}{2pt}
\resizebox{\columnwidth}{!}{%
\setlength{\tabcolsep}{3pt}
\begin{tabular}{c|ccccccc}
\textbf{Method} & \textbf{Graph generator} & \textbf{Edge Existence} & \textbf{Node degree} & \textbf{Node count} & \textbf{Edge count} & \textbf{Connected nodes} & \textbf{Cycle check} \\ \hline
\multirow{7}{*}{\begin{sideways}\method{zero-shot}\end{sideways}}
& \cellcolor{gray!25} Overall &\cellcolor{gray!25} \underline{49.1} &\cellcolor{gray!25} 17.6 &\cellcolor{gray!25} 23.0 &\cellcolor{gray!25} 12.1 &\cellcolor{gray!25} 23.3 &\cellcolor{gray!25} \underline{75.2} \\ 
& ER & 45.1 & 13.6 & 22.1 & 11.7 & 14.9 &  76.3\\
& BA &  50.2 & 18.0 & 24.9 & 13.6 & 20.1 &  72.0\\
& SBM & 45.0 & 13.8 & 21.9 & 9.2 & 13.8 &  86.5\\
& Star & 58.0 & 34.0 & 32.8 & 31.7 & 61.7  & 8.1\\
& SFN & 57.6 & 23.1 & 19.9 & 8.0 & 38.1 & 90.0\\
& Path & \textbf{60.9} & 14.8 & 31.9 & 28.8 & 26.6 & 5.9\\
& Complete & 19.8 & 12.6 & 20.7 & 6.2 & 13.3 & \textbf{91.7}\\
\hline
\multirow{7}{*}{\begin{sideways}\method{cot}\end{sideways}}
& \cellcolor{gray!25} Overall & \cellcolor{gray!25}40.4 & \cellcolor{gray!25}\underline{29.6} &\cellcolor{gray!25} \underline{31.7} &\cellcolor{gray!25} \underline{12.2} &\cellcolor{gray!25} \underline{24.3} &\cellcolor{gray!25} 59.5\\ 
& ER & 41.2 & 28.4 & 28.8 & 12.6 & 12.8 &  61.2\\
& BA & 40.0 & 30.0 & 35.0 & 14.3 & 20.8 &  58.5\\
& SBM & 40.3 & 26.5 & 30.2 & 8.7 & 13.0 &  65.8\\
& Star & 40.3 & \textbf{38.0} & \textbf{41.8} & \textbf{31.6} & \textbf{68.6} & 21.3\\
& SFN & 40.2 & 32.2 & 30.8 & 7.1 & 43.2 &  66.0\\
& Path & 42.0 & 35.1 & 35.3 & 31.1 & 27.6  & 19.7\\
& Complete & 39.6 & 21.9 & 28.9 & 3.9 & 14.6 & 69.3\\
\end{tabular}
}
\caption{Comparing different graph generators on different graph tasks on PaLM 62B. The most effective prompting heuristic is highlighted with an underline, and the top-performing graph generator algorithm for the respective heuristic is highlighted in bold.}
\label{table:graph-generators}
\end{table}

\textbf{Distractive statements in the graph encoding function disrupt the performance of the LLM.} The accuracy of node degree, node count, and connected nodes tasks is highest for star and path graphs. This is likely because the star and path graphs are more likely to have fewer edges and their graph encoding is most likely shorter with less distracting statements to these tasks. This is also evident from the accuracy of these tasks being among the lowest in complete graphs, which have many edges to specify and therefore many distractors. 

\textbf{Adding out-of-distribution few-shot examples helped the LLM.} Similarly to the experiment in Section~\ref{sec:graph-encoders}, adding few-shot examples and their chain of thought in \method{cot} prompting helped on most tasks. The key difference between the few-shot examples in this experiment and the previous one is that in this case, the examples are not required to come from the same graph generator algorithm. This shows that few-shot examples do not need to come from the same generator for the LLM to be helpful, and their main role is to explain the task to the LLM.

\begin{myinsight}
The performance of large language models (LLMs) on graph tasks is significantly impacted by the graph structure and the distracting statements in the graph encoding function. Graphs with fewer edges and less complex encodings tend to perform better on most tasks. Adding few-shot examples, even if they are out-of-distribution, can help the LLM to perform better on most tasks. 
\end{myinsight}

\section{Related Work}

\textbf{In-context learning.}
One common approach for reasoning with LLMs is to pre-train it on a large corpus of text that is closely related to the reasoning task. This has been shown to improve the performance~\citep{hendrycks2021measuring,shen2021generate}, but it can be computationally expensive, especially for larger models.
Additionally, fine-tuning often demands domain-specific data and human expertise, adding to the cost.
\cite{fewshot} has demonstrated the capabilities of LLMs in tackling novel tasks with little or no training data.
The \method{few-shot} method inserts $k$ in-context input-output pairs before the test input and has been shown to significantly improve the performance of the LLM on unseen tasks. Recent research has proposed strategies to improve the selection of in-context demonstrations, such as retrieving semantically similar examples~\citep{liu2021makes}, employing chain-of-thought reasoning~\citep{wei2022chain}, and decomposing tasks into sub-problems using least-to-most prompting~\citep{zhou2022least}.
In this work, we focus on evaluating and enhancing LLMs on basic graph reasoning tasks. 
We exploit some of the ideas in the literature and compare their results.

\textbf{Text-based reasoning with LLMs.}
Numerous models have been proposed for text-based reasoning employing LLMs~(see~\citep{huang2022towards} for a survey).
One approach to reasoning with LLMs is modular reasoning. This methodology divides the problem into smaller modules, utilizing distinct LMs to address each module \citep{zhou2022least, kazemi2022lambada, khot2022decomposed}. Another approach to reasoning with LLMs aims to predict the output of a question in a single LM call. This study primarily focuses on the latter method.

\textbf{Knowledge-Augmented LLMs.} Another body of work is concerned with the use of knowledge (frequently stored in \textit{knowledge graphs} (KGs)) to improve LLM understanding of the world \citep{pan2023unifying}.  Several different methodologies have been proposed which range from generating additional training data from KGs \citep{guu2020retrieval,lewis2020retrieval,agarwal2021knowledge} to extending pretraining  \citep{yasunaga2022deep,jin2023patton}.

\textbf{Reasoning on graphs using LLMs.}
The combination of graph learning and reasoning with LLMs is a rapidly growing area.
InstructGLM~\citep{ye2023natural} proposed an instruction-finetuned LLM for performing node classification. \cite{chen2023exploring} used LLMs as enhancers to exploit text attributes to be used in a graph learning model or as predictors for node classification on text-attributed graphs. The closest work to ours is \cite{wang2023can}, which proposed a set of tasks for benchmarking LLMs on graphs. However, this work omitted several natural graph tasks, lacked variety in the type of graph structure considered, and fixed the graph and question encoder function. They conclude that LLMs have preliminary graph reasoning abilities on somewhat complex graph tasks.

\textbf{Present work.} In this study, we focus on basic graph tasks, which are essential intermediate steps for more complex reasoning tasks on graphs. We conduct extensive experiments on graph and question encoder functions, as well as a wide range of graph generator functions. We provide an extensive study of graph encoding methods for black-box LLM usage, and introduce \gqa, a new graph benchmark that illustrates the effect of graph structure on LLM encoding. We also provide insights and best practices for encoding graphs as text for use in LLMs.

\section{Conclusions}

In this work, we have presented the first comprehensive study of encoding graph-structured data as text for consumption by LLMs.
We show that LLM performance on graph reasoning tasks varies on three fundamental levels: (1) the graph encoding method, (2) the nature of the graph task itself, and (3) interestingly, the very structure of the graph considered.
These novel results provide valuable insight on strategies for encoding graphs as text -- which can boost performance on graph reasoning tasks inside LLMs by 4.8\% to 61.8\%.
We believe that this is a fruitful avenue for further investigation, and hope that our \gqa{} benchmark tasks inspire additional work in the area.

\bibliography{iclr2024_conference}

\begin{thebibliography}{47}
\providecommand{\natexlab}[1]{#1}
\providecommand{\url}[1]{\texttt{#1}}
\expandafter\ifx\csname urlstyle\endcsname\relax
  \providecommand{\doi}[1]{doi: #1}\else
  \providecommand{\doi}{doi: \begingroup \urlstyle{rm}\Url}\fi

\bibitem[Agarwal et~al.(2021)Agarwal, Ge, Shakeri, and
  Al-Rfou]{agarwal2021knowledge}
Oshin Agarwal, Heming Ge, Siamak Shakeri, and Rami Al-Rfou.
\newblock Knowledge graph based synthetic corpus generation for
  knowledge-enhanced language model pre-training, 2021.

\bibitem[Albert \& Barab{\'a}si(2002)Albert and Barab{\'a}si]{barabasi}
R{\'e}ka Albert and Albert-L{\'a}szl{\'o} Barab{\'a}si.
\newblock Statistical mechanics of complex networks.
\newblock \emph{Reviews of modern physics}, 74\penalty0 (1):\penalty0 47, 2002.

\bibitem[Anil et~al.(2023)Anil, Dai, Firat, Johnson, Lepikhin, Passos, Shakeri,
  Taropa, Bailey, Chen, et~al.]{anil2023palm}
Rohan Anil, Andrew~M Dai, Orhan Firat, Melvin Johnson, Dmitry Lepikhin,
  Alexandre Passos, Siamak Shakeri, Emanuel Taropa, Paige Bailey, Zhifeng Chen,
  et~al.
\newblock Palm 2 technical report.
\newblock \emph{arXiv preprint arXiv:2305.10403}, 2023.

\bibitem[Barab{\'a}si \& Albert(1999)Barab{\'a}si and Albert]{scalefree}
Albert-L{\'a}szl{\'o} Barab{\'a}si and R{\'e}ka Albert.
\newblock Emergence of scaling in random networks.
\newblock \emph{science}, 286\penalty0 (5439):\penalty0 509--512, 1999.

\bibitem[Brown et~al.(2020{\natexlab{a}})Brown, Mann, Ryder, Subbiah, Kaplan,
  Dhariwal, Neelakantan, Shyam, Sastry, Askell, et~al.]{brown2020language}
Tom Brown, Benjamin Mann, Nick Ryder, Melanie Subbiah, Jared~D Kaplan, Prafulla
  Dhariwal, Arvind Neelakantan, Pranav Shyam, Girish Sastry, Amanda Askell,
  et~al.
\newblock Language models are few-shot learners.
\newblock \emph{Advances in neural information processing systems},
  33:\penalty0 1877--1901, 2020{\natexlab{a}}.

\bibitem[Brown et~al.(2020{\natexlab{b}})Brown, Mann, Ryder, Subbiah, Kaplan,
  Dhariwal, Neelakantan, Shyam, Sastry, Askell, et~al.]{fewshot}
Tom Brown, Benjamin Mann, Nick Ryder, Melanie Subbiah, Jared~D Kaplan, Prafulla
  Dhariwal, Arvind Neelakantan, Pranav Shyam, Girish Sastry, Amanda Askell,
  et~al.
\newblock Language models are few-shot learners.
\newblock \emph{Advances in neural information processing systems},
  33:\penalty0 1877--1901, 2020{\natexlab{b}}.

\bibitem[Bubeck et~al.(2023)Bubeck, Chandrasekaran, Eldan, Gehrke, Horvitz,
  Kamar, Lee, Lee, Li, Lundberg, et~al.]{bubeck2023sparks}
S{\'e}bastien Bubeck, Varun Chandrasekaran, Ronen Eldan, Johannes Gehrke, Eric
  Horvitz, Ece Kamar, Peter Lee, Yin~Tat Lee, Yuanzhi Li, Scott Lundberg,
  et~al.
\newblock Sparks of artificial general intelligence: Early experiments with
  gpt-4.
\newblock \emph{arXiv preprint arXiv:2303.12712}, 2023.

\bibitem[Chen et~al.(2023)Chen, Mao, Li, Jin, Wen, Wei, Wang, Yin, Fan, Liu,
  et~al.]{chen2023exploring}
Zhikai Chen, Haitao Mao, Hang Li, Wei Jin, Hongzhi Wen, Xiaochi Wei, Shuaiqiang
  Wang, Dawei Yin, Wenqi Fan, Hui Liu, et~al.
\newblock Exploring the potential of large language models (llms) in learning
  on graphs.
\newblock \emph{arXiv preprint arXiv:2307.03393}, 2023.

\bibitem[Chowdhery et~al.(2022)Chowdhery, Narang, Devlin, Bosma, Mishra,
  Roberts, Barham, Chung, Sutton, Gehrmann, et~al.]{chowdhery2022palm}
Aakanksha Chowdhery, Sharan Narang, Jacob Devlin, Maarten Bosma, Gaurav Mishra,
  Adam Roberts, Paul Barham, Hyung~Won Chung, Charles Sutton, Sebastian
  Gehrmann, et~al.
\newblock Palm: Scaling language modeling with pathways.
\newblock \emph{arXiv preprint arXiv:2204.02311}, 2022.

\bibitem[Chung et~al.(2022)Chung, Hou, Longpre, Zoph, Tay, Fedus, Li, Wang,
  Dehghani, Brahma, et~al.]{flan}
Hyung~Won Chung, Le~Hou, Shayne Longpre, Barret Zoph, Yi~Tay, William Fedus,
  Eric Li, Xuezhi Wang, Mostafa Dehghani, Siddhartha Brahma, et~al.
\newblock Scaling instruction-finetuned language models.
\newblock \emph{arXiv preprint arXiv:2210.11416}, 2022.

\bibitem[Clark et~al.(2020)Clark, Tafjord, and
  Richardson]{clark2020transformers}
Peter Clark, Oyvind Tafjord, and Kyle Richardson.
\newblock Transformers as soft reasoners over language.
\newblock \emph{arXiv preprint arXiv:2002.05867}, 2020.

\bibitem[Devlin et~al.(2018)Devlin, Chang, Lee, and Toutanova]{devlin2018bert}
Jacob Devlin, Ming-Wei Chang, Kenton Lee, and Kristina Toutanova.
\newblock Bert: Pre-training of deep bidirectional transformers for language
  understanding.
\newblock \emph{arXiv preprint arXiv:1810.04805}, 2018.

\bibitem[Dwivedi \& Bresson(2020)Dwivedi and
  Bresson]{dwivedi2020generalization}
Vijay~Prakash Dwivedi and Xavier Bresson.
\newblock A generalization of transformer networks to graphs.
\newblock \emph{arXiv preprint arXiv:2012.09699}, 2020.

\bibitem[Erd\H{o}s \& R\'enyi(1959)Erd\H{o}s and R\'enyi]{erdds1959random}
Paul Erd\H{o}s and Alfred R\'enyi.
\newblock On random graphs.
\newblock \emph{Publicationes Mathematicae Debrecen}, 6:\penalty0 290--297,
  1959.

\bibitem[Guu et~al.(2020)Guu, Lee, Tung, Pasupat, and Chang]{guu2020retrieval}
Kelvin Guu, Kenton Lee, Zora Tung, Panupong Pasupat, and Mingwei Chang.
\newblock Retrieval augmented language model pre-training.
\newblock In \emph{International conference on machine learning}, pp.\
  3929--3938. PMLR, 2020.

\bibitem[Hagberg et~al.(2008)Hagberg, Swart, and S~Chult]{networkx}
Aric Hagberg, Pieter Swart, and Daniel S~Chult.
\newblock Exploring network structure, dynamics, and function using networkx.
\newblock Technical report, Los Alamos National Lab.(LANL), Los Alamos, NM
  (United States), 2008.

\bibitem[Hendrycks et~al.(2021)Hendrycks, Burns, Kadavath, Arora, Basart, Tang,
  Song, and Steinhardt]{hendrycks2021measuring}
Dan Hendrycks, Collin Burns, Saurav Kadavath, Akul Arora, Steven Basart, Eric
  Tang, Dawn Song, and Jacob Steinhardt.
\newblock Measuring mathematical problem solving with the math dataset.
\newblock \emph{arXiv preprint arXiv:2103.03874}, 2021.

\bibitem[Holland et~al.(1983)Holland, Laskey, and
  Leinhardt]{holland1983stochastic}
Paul~W Holland, Kathryn~Blackmond Laskey, and Samuel Leinhardt.
\newblock Stochastic blockmodels: First steps.
\newblock \emph{Social networks}, 5\penalty0 (2):\penalty0 109--137, 1983.

\bibitem[Hu et~al.(2021)Hu, Shen, Wallis, Allen-Zhu, Li, Wang, Wang, and
  Chen]{hu2021lora}
Edward~J Hu, Yelong Shen, Phillip Wallis, Zeyuan Allen-Zhu, Yuanzhi Li, Shean
  Wang, Lu~Wang, and Weizhu Chen.
\newblock Lora: Low-rank adaptation of large language models.
\newblock \emph{arXiv preprint arXiv:2106.09685}, 2021.

\bibitem[Huang \& Chang(2022)Huang and Chang]{huang2022towards}
Jie Huang and Kevin Chen-Chuan Chang.
\newblock Towards reasoning in large language models: A survey.
\newblock \emph{arXiv preprint arXiv:2212.10403}, 2022.

\bibitem[Jin et~al.(2023)Jin, Zhang, Zhang, Meng, Zhang, Zhu, and
  Han]{jin2023patton}
Bowen Jin, Wentao Zhang, Yu~Zhang, Yu~Meng, Xinyang Zhang, Qi~Zhu, and Jiawei
  Han.
\newblock Patton: Language model pretraining on text-rich networks.
\newblock \emph{arXiv preprint arXiv:2305.12268}, 2023.

\bibitem[Kazemi et~al.(2022)Kazemi, Kim, Bhatia, Xu, and
  Ramachandran]{kazemi2022lambada}
Seyed~Mehran Kazemi, Najoung Kim, Deepti Bhatia, Xin Xu, and Deepak
  Ramachandran.
\newblock Lambada: Backward chaining for automated reasoning in natural
  language.
\newblock \emph{arXiv preprint arXiv:2212.13894}, 2022.

\bibitem[Khot et~al.(2022)Khot, Trivedi, Finlayson, Fu, Richardson, Clark, and
  Sabharwal]{khot2022decomposed}
Tushar Khot, Harsh Trivedi, Matthew Finlayson, Yao Fu, Kyle Richardson, Peter
  Clark, and Ashish Sabharwal.
\newblock Decomposed prompting: A modular approach for solving complex tasks.
\newblock \emph{arXiv preprint arXiv:2210.02406}, 2022.

\bibitem[Kojima et~al.(2022)Kojima, Gu, Reid, Matsuo, and
  Iwasawa]{kojima2022large}
Takeshi Kojima, Shixiang~Shane Gu, Machel Reid, Yutaka Matsuo, and Yusuke
  Iwasawa.
\newblock Large language models are zero-shot reasoners.
\newblock \emph{Advances in neural information processing systems},
  35:\penalty0 22199--22213, 2022.

\bibitem[Lester et~al.(2021)Lester, Al-Rfou, and Constant]{lester2021power}
Brian Lester, Rami Al-Rfou, and Noah Constant.
\newblock The power of scale for parameter-efficient prompt tuning.
\newblock \emph{arXiv preprint arXiv:2104.08691}, 2021.

\bibitem[Lewis et~al.(2020)Lewis, Perez, Piktus, Petroni, Karpukhin, Goyal,
  K{\"u}ttler, Lewis, Yih, Rockt{\"a}schel, et~al.]{lewis2020retrieval}
Patrick Lewis, Ethan Perez, Aleksandra Piktus, Fabio Petroni, Vladimir
  Karpukhin, Naman Goyal, Heinrich K{\"u}ttler, Mike Lewis, Wen-tau Yih, Tim
  Rockt{\"a}schel, et~al.
\newblock Retrieval-augmented generation for knowledge-intensive nlp tasks.
\newblock \emph{Advances in Neural Information Processing Systems},
  33:\penalty0 9459--9474, 2020.

\bibitem[Liu et~al.(2021)Liu, Shen, Zhang, Dolan, Carin, and
  Chen]{liu2021makes}
Jiachang Liu, Dinghan Shen, Yizhe Zhang, Bill Dolan, Lawrence Carin, and Weizhu
  Chen.
\newblock What makes good in-context examples for gpt-$3 $?
\newblock \emph{arXiv preprint arXiv:2101.06804}, 2021.

\bibitem[M{\"u}ller et~al.(2023)M{\"u}ller, Galkin, Morris, and
  Ramp{\'a}{\v{s}}ek]{muller2023attending}
Luis M{\"u}ller, Mikhail Galkin, Christopher Morris, and Ladislav
  Ramp{\'a}{\v{s}}ek.
\newblock Attending to graph transformers.
\newblock \emph{arXiv preprint arXiv:2302.04181}, 2023.

\bibitem[Ouyang et~al.(2022)Ouyang, Wu, Jiang, Almeida, Wainwright, Mishkin,
  Zhang, Agarwal, Slama, Ray, et~al.]{ouyang2022training}
Long Ouyang, Jeffrey Wu, Xu~Jiang, Diogo Almeida, Carroll Wainwright, Pamela
  Mishkin, Chong Zhang, Sandhini Agarwal, Katarina Slama, Alex Ray, et~al.
\newblock Training language models to follow instructions with human feedback.
\newblock \emph{Advances in Neural Information Processing Systems},
  35:\penalty0 27730--27744, 2022.

\bibitem[Palowitch et~al.(2022)Palowitch, Tsitsulin, Mayer, and
  Perozzi]{palowitch2022graphworld}
John Palowitch, Anton Tsitsulin, Brandon Mayer, and Bryan Perozzi.
\newblock Graphworld: Fake graphs bring real insights for gnns.
\newblock In \emph{Proceedings of the 28th ACM SIGKDD Conference on Knowledge
  Discovery and Data Mining}, pp.\  3691--3701, 2022.

\bibitem[Pan et~al.(2023)Pan, Luo, Wang, Chen, Wang, and Wu]{pan2023unifying}
Shirui Pan, Linhao Luo, Yufei Wang, Chen Chen, Jiapu Wang, and Xindong Wu.
\newblock Unifying large language models and knowledge graphs: A roadmap, 2023.

\bibitem[Pryzant et~al.(2023)Pryzant, Iter, Li, Lee, Zhu, and
  Zeng]{pryzant2023automatic}
Reid Pryzant, Dan Iter, Jerry Li, Yin~Tat Lee, Chenguang Zhu, and Michael Zeng.
\newblock Automatic prompt optimization with" gradient descent" and beam
  search.
\newblock \emph{arXiv preprint arXiv:2305.03495}, 2023.

\bibitem[Schneider et~al.(2022)Schneider, Schopf, Vladika, Galkin, Simperl, and
  Matthes]{schneider2022decade}
Phillip Schneider, Tim Schopf, Juraj Vladika, Mikhail Galkin, Elena Simperl,
  and Florian Matthes.
\newblock A decade of knowledge graphs in natural language processing: A
  survey.
\newblock \emph{arXiv preprint arXiv:2210.00105}, 2022.

\bibitem[Shen et~al.(2021)Shen, Yin, Li, Shang, Jiang, Zhang, and
  Liu]{shen2021generate}
Jianhao Shen, Yichun Yin, Lin Li, Lifeng Shang, Xin Jiang, Ming Zhang, and Qun
  Liu.
\newblock Generate \& rank: A multi-task framework for math word problems.
\newblock \emph{arXiv preprint arXiv:2109.03034}, 2021.

\bibitem[Vaswani et~al.(2017)Vaswani, Shazeer, Parmar, Uszkoreit, Jones, Gomez,
  Kaiser, and Polosukhin]{vaswani2017attention}
Ashish Vaswani, Noam Shazeer, Niki Parmar, Jakob Uszkoreit, Llion Jones,
  Aidan~N Gomez, {\L}ukasz Kaiser, and Illia Polosukhin.
\newblock Attention is all you need.
\newblock \emph{Advances in neural information processing systems}, 30, 2017.

\bibitem[Wang et~al.(2023)Wang, Feng, He, Tan, Han, and Tsvetkov]{wang2023can}
Heng Wang, Shangbin Feng, Tianxing He, Zhaoxuan Tan, Xiaochuang Han, and Yulia
  Tsvetkov.
\newblock Can language models solve graph problems in natural language?
\newblock \emph{arXiv preprint arXiv:2305.10037}, 2023.

\bibitem[Wei et~al.(2022)Wei, Wang, Schuurmans, Bosma, Xia, Chi, Le, Zhou,
  et~al.]{wei2022chain}
Jason Wei, Xuezhi Wang, Dale Schuurmans, Maarten Bosma, Fei Xia, Ed~Chi, Quoc~V
  Le, Denny Zhou, et~al.
\newblock Chain-of-thought prompting elicits reasoning in large language
  models.
\newblock \emph{Advances in Neural Information Processing Systems},
  35:\penalty0 24824--24837, 2022.

\bibitem[Yang et~al.(2023)Yang, Wang, Lu, Liu, Le, Zhou, and
  Chen]{yang2023large}
Chengrun Yang, Xuezhi Wang, Yifeng Lu, Hanxiao Liu, Quoc~V Le, Denny Zhou, and
  Xinyun Chen.
\newblock Large language models as optimizers.
\newblock \emph{arXiv preprint arXiv:2309.03409}, 2023.

\bibitem[Yasir et~al.(2023)Yasir, Palowitch, Tsitsulin, Tran-Thanh, and
  Perozzi]{yasir2023examining}
Mustafa Yasir, John Palowitch, Anton Tsitsulin, Long Tran-Thanh, and Bryan
  Perozzi.
\newblock Examining the effects of degree distribution and homophily in graph
  learning models, 2023.

\bibitem[Yasunaga et~al.(2022)Yasunaga, Bosselut, Ren, Zhang, Manning, Liang,
  and Leskovec]{yasunaga2022deep}
Michihiro Yasunaga, Antoine Bosselut, Hongyu Ren, Xikun Zhang, Christopher~D
  Manning, Percy Liang, and Jure Leskovec.
\newblock Deep bidirectional language-knowledge graph pretraining, 2022.

\bibitem[Ye et~al.(2023)Ye, Zhang, Wang, Xu, and Zhang]{ye2023natural}
Ruosong Ye, Caiqi Zhang, Runhui Wang, Shuyuan Xu, and Yongfeng Zhang.
\newblock Natural language is all a graph needs.
\newblock \emph{arXiv preprint arXiv:2308.07134}, 2023.

\bibitem[Zhang et~al.(2020)Zhang, Zhang, Xia, and Sun]{zhang2020graph}
Jiawei Zhang, Haopeng Zhang, Congying Xia, and Li~Sun.
\newblock Graph-bert: Only attention is needed for learning graph
  representations.
\newblock \emph{arXiv preprint arXiv:2001.05140}, 2020.

\bibitem[Zhang et~al.(2023{\natexlab{a}})Zhang, Florin, Lee, Niknafs,
  Marginean, Wang, Tyser, Chin, Hicke, Singh, et~al.]{zhang2023exploring}
Sarah~J Zhang, Samuel Florin, Ariel~N Lee, Eamon Niknafs, Andrei Marginean,
  Annie Wang, Keith Tyser, Zad Chin, Yann Hicke, Nikhil Singh, et~al.
\newblock Exploring the mit mathematics and eecs curriculum using large
  language models.
\newblock \emph{arXiv preprint arXiv:2306.08997}, 2023{\natexlab{a}}.

\bibitem[Zhang et~al.(2023{\natexlab{b}})Zhang, Li, Cui, Cai, Liu, Fu, Huang,
  Zhao, Zhang, Chen, et~al.]{zhang2023siren}
Yue Zhang, Yafu Li, Leyang Cui, Deng Cai, Lemao Liu, Tingchen Fu, Xinting
  Huang, Enbo Zhao, Yu~Zhang, Yulong Chen, et~al.
\newblock Siren's song in the ai ocean: A survey on hallucination in large
  language models.
\newblock \emph{arXiv preprint arXiv:2309.01219}, 2023{\natexlab{b}}.

\bibitem[Zhao et~al.(2023)Zhao, Zhou, Li, Tang, Wang, Hou, Min, Zhang, Zhang,
  Dong, et~al.]{zhao2023survey}
Wayne~Xin Zhao, Kun Zhou, Junyi Li, Tianyi Tang, Xiaolei Wang, Yupeng Hou,
  Yingqian Min, Beichen Zhang, Junjie Zhang, Zican Dong, et~al.
\newblock A survey of large language models.
\newblock \emph{arXiv preprint arXiv:2303.18223}, 2023.

\bibitem[Zhou et~al.(2022{\natexlab{a}})Zhou, Sch{\"a}rli, Hou, Wei, Scales,
  Wang, Schuurmans, Cui, Bousquet, Le, et~al.]{zhou2022least}
Denny Zhou, Nathanael Sch{\"a}rli, Le~Hou, Jason Wei, Nathan Scales, Xuezhi
  Wang, Dale Schuurmans, Claire Cui, Olivier Bousquet, Quoc Le, et~al.
\newblock Least-to-most prompting enables complex reasoning in large language
  models.
\newblock \emph{arXiv preprint arXiv:2205.10625}, 2022{\natexlab{a}}.

\bibitem[Zhou et~al.(2022{\natexlab{b}})Zhou, Muresanu, Han, Paster, Pitis,
  Chan, and Ba]{zhou2022large}
Yongchao Zhou, Andrei~Ioan Muresanu, Ziwen Han, Keiran Paster, Silviu Pitis,
  Harris Chan, and Jimmy Ba.
\newblock Large language models are human-level prompt engineers.
\newblock \emph{arXiv preprint arXiv:2211.01910}, 2022{\natexlab{b}}.

\end{thebibliography}
\bibliographystyle{iclr2024_conference}

\newpage
\appendix
\section{Appendix}

\subsection{Graph Encoding Function}\label{sec:graph-encoding-detail}

We conducted an investigation into various methodologies for representing graphs as text. This process of encoding graphs as text can be separated into two key inquiries:
First, the encoding of nodes within the graph, and second the encoding of edges between the nodes.

\textit{Encoding Nodes}.
Regarding the encoding of nodes, we examined several techniques, including:
\begin{itemize}
    \item Integer encoding (\eg \textit{Node 0}).
    \item Utilizing well-known English first names~(\eg David).
    \item Utilizing popular character names in television series Game of Thrones and South Park.
    \item Incorporating the first names of American politicians.
    \item Employing alphabet letters for representation.
\end{itemize}

\textit{Representing Edges}.
Regarding the encoding of the edges, we examined the following techniques:
\begin{itemize}
    \item Parenthesis: describing edges as (source node, target node).
    \item Friendship: source node and target node are friends.
    \item Coauthorship: source node and target node wrote a paper together.
    \item Social network: source node and target node are connected.
    \item Arrows: source node $\xrightarrow[]{}$ target node.
    \item Incident: source node is connected to target nodes.
\end{itemize}

Combining the node and edge encoding, we start with the following list of graph encoding functions:
\begin{itemize}

    \item{\textbf{Adjacency.}} using integer node encoding and parenthesis edge encoding.
    \item{\textbf{Incident.}} using integer node encoding and incident edge encoding.
    \item{\textbf{Friendship}.} using well-known english first names as node encoding and friendship edge encoding.
    \item{\textbf{Co-authorship.}} using well-known english first names as node encoding and coauthorship edge encoding.
    \item{\textbf{SP.}} using South Park character names as node encoding and friendship as edge encoding.
    \item{\textbf{GOT.}} using Game of Thrones character names as node encoding and friendship as edge encoding.
    \item{\textbf{Social network.}} using well-known English first names and social network edge encoding.
    \item{\textbf{Politician.}} using politician American politician first names and social network edge encoding.
    \item{\textbf{Expert.}} employing alphabet letters for node encoding and arrows as edge encoding. The encoding starts with ``You are a graph analyst'' (expert prompting~\citep{zhang2023exploring}). 
\end{itemize}

\begin{figure}[t]
  \centering
  \includegraphics[width=0.3\textwidth]{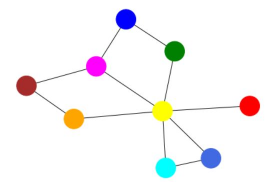}
  \caption{Running example graph for all graph encoding functions.
    }\label{fig:example_graph}
\end{figure}

Here, we provide the full details for the graph encoding functions for the graph example in \Cref{fig:example_graph}.

\noindent\fbox{%
\parbox{\textwidth}{
    \textbf{Adjacency:} 
In an undirected graph, (i,j) means that node i and node j are connected with an undirected edge. G describes a graph among nodes 0, 1, 2, 3, 4, 5, 6, 7, and 8.\\
The edges in G are: (0, 1) (0, 2) (1, 2) (2, 3) (2, 4) (2, 5) (2, 7) (3, 8) (5, 6) (6, 7) (7, 8).
}
}

\noindent\fbox{%
\parbox{\textwidth}{
\textbf{Incident:} 
G describes a graph among 0, 1, 2, 3, 4, 5, 6, 7, and 8.\\
In this graph:\\
Node 0 is connected to nodes 1, 2.\\
Node 1 is connected to nodes 0, 2.\\
Node 2 is connected to nodes 0, 1, 3, 4, 5, 7.\\
Node 3 is connected to nodes 2, 8.\\
Node 4 is connected to node 2.\\
Node 5 is connected to nodes 2, 6.\\
Node 6 is connected to nodes 7, 5.\\
Node 7 is connected to nodes 2, 8, 6.\\
Node 8 is connected to nodes 3, 7.
}
}

\noindent\fbox{%
\parbox{\textwidth}{
\textbf{Co-authorship:} 
G describes a co-authorship graph among James, Robert, John, Michael, David, Mary, Patricia, Jennifer, and Linda.\\
In this co-authorship graph:\\
James and Robert wrote a paper together.\\
James and John wrote a paper together.\\
Robert and John wrote a paper together.\\
John and Michael wrote a paper together.\\
John and David wrote a paper together.\\
John and Mary wrote a paper together.\\
John and Jennifer wrote a paper together.\\
Michael and Linda wrote a paper together.\\
Mary and Patricia wrote a paper together.\\
Patricia and Jennifer wrote a paper together.\\
Jennifer and Linda wrote a paper together.
}
}

\noindent\fbox{%
\parbox{\textwidth}{
\textbf{Friendship:} 
G describes a friendship graph among James, Robert, John, Michael, David, Mary, Patricia, Jennifer, and Linda.\\
We have the following edges in G:\\
James and Robert are friends.\\
James and John are friends.\\
Robert and John are friends.\\
John and Michael are friends.\\
John and David are friends.\\
John and Mary are friends.\\
John and Jennifer are friends.\\
Michael and Linda are friends.\\
Mary and Patricia are friends.\\
Patricia and Jennifer are friends.\\
Jennifer and Linda are friends.
}
}

\noindent\fbox{%
\parbox{\textwidth}{
\textbf{SP:} 
G describes a friendship graph among Eric, Kenny, Kyle, Stan, Tolkien, Heidi, Bebe, Liane, and Sharon.\\
In this friendship graph: \\
Eric and Kenny are friends, Eric and Kyle are friends, Kenny and Kyle are friends, Kyle and Stan are friends, Kyle and Tolkien are friends, Kyle and Heidi are friends, Kyle and Liane are friends, Stan and Sharon are friends, Heidi and Bebe are friends, Bebe and Liane are friends, Liane and Sharon are friends.
}
}

\noindent\fbox{%
\parbox{\textwidth}{
\textbf{GOT:} 
G describes a friendship graph among Ned, Cat, Daenerys, Jon, Bran, Sansa, Arya, Cersei, and Jaime.\\
In this friendship graph: Ned and Cat are friends, Ned and Daenerys are friends, Cat and Daenerys are friends, Daenerys and Jon are friends, Daenerys and Bran are friends, Daenerys and Sansa are friends, Daenerys and Cersei are friends, Jon and Jaime are friends, Sansa and Arya are friends, Arya and Cersei are friends, Cersei and Jaime are friends.
}
}

\noindent\fbox{%
\parbox{\textwidth}{
\textbf{Social Network:} 
G describes a social network graph among James, Robert, John, Michael, David, Mary, Patricia, Jennifer, and Linda.\\
We have the following edges in G:\\
James and Robert are connected.\\
James and John are connected.\\
Robert and John are connected.\\
John and Michael are connected.\\
John and David are connected.\\
John and Mary are connected.\\
John and Jennifer are connected.\\
Michael and Linda are connected.\\
Mary and Patricia are connected.\\
Patricia and Jennifer are connected.\\
Jennifer and Linda are connected.
}
}

\noindent\fbox{%
\parbox{\textwidth}{
\textbf{Politician:} 
G describes a social network graph among Barack, Jimmy, Arnold, Bernie, Bill, Kamala, Hillary, Elizabeth, and John.\\
We have the following edges in G:\\
Barack and Jimmy are connected.\\
Barack and Arnold are connected.\\
Jimmy and Arnold are connected.\\
Arnold and Bernie are connected.\\
Arnold and Bill are connected.\\
Arnold and Kamala are connected.\\
Arnold and Elizabeth are connected.\\
Bernie and John are connected.\\
Kamala and Hillary are connected.\\
Hillary and Elizabeth are connected.\\
Elizabeth and John are connected.
}
}

\noindent\fbox{%
\parbox{\textwidth}{
\textbf{Expert:} 
You are a graph analyst and you have been given a graph G among A, B, C, D, E, F, G, H, and I.
G has the following undirected edges:\\
A -$>$ B\\
A -$>$ C\\
B -$>$ C\\
C -$>$ D\\
C -$>$ E\\
C -$>$ F\\
C -$>$ H\\
D -$>$ I\\
F -$>$ G\\
G -$>$ H\\
H -$>$ I
}
}

\subsection{Graph Tasks}\label{sec:basic-graph-tasks}
\gqa{} consists of a diverse set of basic graph problems, including:

\begin{itemize}
    \item{\textbf{Edge existence.}} Determine whether a given edge exists in a graph.
    \item{\textbf{Node degree.}} Calculate the degree of a given node in a graph.
    \item{\textbf{Node count.}} Count the number of nodes in a graph.
    \item{\textbf{Edge count.}} Count the number of edges in a graph.
    \item{\textbf{Connected nodes.}} Find all the nodes that are connected to a given node in a graph.
    \item{\textbf{Cycle check.}} Determine whether a graph contains a cycle.
    \item{\textbf{Disconnected nodes.}} Find all the nodes that are not connected to a given node in a graph.
\end{itemize}

These tasks are all relatively simple, but they require LLMs to be able to reason about the relationships between nodes and edges in a graph. 
While adhering to basic graph tasks, we aimed for a diverse set of tasks, including discriminative (\eg cycle check) and generative (\eg connected or disconnected nodes) challenges. These tasks covered various aspects of graph analysis, from existence checks (\eg edge existence) to quantitative assessments (\eg node count), path analysis (\eg cycle check), recall-based tasks (\eg connected nodes), and null space exploration (\eg disconnected nodes).

The basic graph tasks listed above are all essential intermediate steps for more complex reasoning tasks on graphs. For example, to determine the shortest path between two nodes in a graph, we must first be able to find all the nodes that are connected to a given node. To detect communities in a graph, we must first be able to identify all the cycles in the graph. To find the most influential node in a graph, we must first be able to calculate the degree of each node.
These tasks are essential building blocks for more complex reasoning tasks on graphs. 

\subsection{graph encoding Rankings}
\label{sec:encoding_ranking}
To provide recommendations about the best graph encoding function to use for each prompt type, we rank the encoders by their average standing~(in rank order) on each graph task.
The results are presented in \Cref{table:encoder_ranking}, where a lower number is better~(the encoder ranked higher on average).
We note that for most prompting methods, incident encoding performed the best. However, for \method{zero-shot} graph prompting, node tokens with more established representations~(such as politicians or popular fantasy characters) outperformed incident encoding.

\begin{table}[h!]
\begin{tabular}{lccccccc}
\toprule
 graph encoding &  \method{zero-shot} &  \method{zero-cot} &  \method{few-shot} &       \method{cot} &   \method{cot-bag} \\
\midrule
Adjacency &   4.83 &  3.25 &  2.16 &  3.00 &  1.83 \\
Incident &   6.16 &  \textbf{2.58} &  \textbf{2.00} &  \textbf{2.33} &  \textbf{1.33} \\
Co-authorship &   6.08 &  6.33 &  5.58 &  6.75 &  8.83 \\
Friendship &   5.16 &  6.41 &  6.25 &  4.66 &  6.00 \\
SP &   5.16 &  4.50 &  5.25 &  5.75 &  4.66 \\
GOT &   4.33 &  4.08 &  5.83 &  5.00 &  6.25 \\
Social Network &   4.58 &  6.50 &  5.83 &  6.16 &  6.41 \\
Politician &   \textbf{3.50} &  6.33 &  6.25 &  5.58 &  4.00 \\
Expert &   5.16 &  5.00 &  5.83 &  5.75 &  5.66 \\
\bottomrule
\end{tabular}
\caption{Ranking of graph encodings from experiment in \Cref{sec:graph-encoders} (lower better).}
\label{table:encoder_ranking}
\end{table}

\subsection{Implementation Details}\label{sec:implementation-detail}
For our experiments, we used PaLM 62B and PaLM 2 (various sizes) served on a $4 \times 4$ TPU v4 architecture. The decoding temperature was set to zero. 
We used the NetworkX library~\citep{networkx} to generate the random graphs and to find the answers to the graph tasks. 
To generate random graphs, we use Erd\H{o}s-R\'enyi~(ER) graphs~\citep{erdds1959random}, scale-free networks~(SFN)~\citep{scalefree}, Barabási–Albert model~(BA)~\citep{barabasi}, and stochastic block model~(SBM)~\citep{holland1983stochastic}, in addition to star, path, and complete graph generators. 
To generate graphs, we sampled $500$ graphs for each of the following algorithms: ER, BA, SFN, and SBM. We sampled $100$ graphs for path, complete, and star graphs, as these have less variety. All graphs had between $5$ and $20$ nodes. For ER graphs, we sampled the probability for edge creation from $[0, 1]$. For SBM graphs, number of communities has been sampled from $2$ to $10$. 
We are committed to open-sourcing both our code and data upon the acceptance of our paper.

\subsection{Evaluating More LLMs for Graph Tasks with Different Graph Encoding Functions} 

We compared different graph encoding functions on a PaLM 62B~\citep{chowdhery2022palm} in \Cref{sec:graph-encoders}. Here, we provide the results of the same experiment on PaLM 2 XXS, XS, S, and L~\citep{anil2023palm} in \Cref{table:graph-encoders-palm2xxs,table:graph-encoders-palm2xs,table:graph-encoders-palm2s,table:graph-encoders-palm2l}. We also provide results for some instruction-finetuned Flan~\citep{flan} checkpoints of the same models in \Cref{table:graph-encoders-palm2xxsf,table:graph-encoders-palm2xsf}.

\begin{table}[b!]
\centering
\footnotesize
\setlength{\tabcolsep}{2pt}
\resizebox{\columnwidth}{!}{%
\setlength{\tabcolsep}{3pt}
\begin{tabular}{c|c|c|c|c|c|c|c}
Method & Encoding function & Edge Existence & Node degree & Node count & Edge count & Connected nodes & Cycle check \\ \hline
\multirow{9}{*}{\begin{sideways}\method{zero-shot}\end{sideways}}
& Overall & 47.2 & 11.3 & 8.7 & 6.4 & 7.2  & 61.5 \\
\cline{2-8}
& Adjacency & 48.4 & 14.4 & 6.2 & 4.0 & 17.6 & 82.6 \\
& Incident & 45.2 & 13.4 & 7.2 & 5.2 & 11.2 & 68.4 \\
& Co-authorship & 45.4 & 10.8 & 7.4 & 4.6 & 5.2 & 66.4 \\
& Friendship & 42.8 & 10.8 & 5.4 & 5.6 & 1.6 & 65.2 \\
& SP & 56.6 & 11.0 & 7.2 & 5.8 & 3.0 & 26.6 \\
& GOT & 56.4 & 7.8 & 6.0 & 7.0 & 2.0 & 51.8 \\
& Social Network & 51.2 & 11.0 & 7.8 & 5.4 & 5.2 & 74.4 \\
& Politician & 40.6 & 12.0 & 9.4 & 6.8 & 10.0 & 73.2 \\
& Expert & 38.0 & 10.4 & 21.4 & 12.8 & 9.0 & 45.2 \\
\hline
\multirow{9}{*}{\begin{sideways}\method{zero-cot}\end{sideways}}
& Overall & 34.1 & 8.6 & 1.7 & 2.2 & 6.0 & 13.7 \\
\cline{2-8}
& Adjacency & 20.2 & 19.0 & 2.4 & 2.0 & 9.0 & 16.4 \\
& Incident & 45.0 & 36.0 & 1.2 & 6.0 & 16.0 & 37.8 \\
& Co-authorship & 48.8 & 22.0 & 0.8 & 4.4 & 11.8 & 31.6 \\
& Friendship & 43.2 & 0.2 & 0.6 & 1.6 & 2.0 &  10.8 \\
& SP & 30.8 & 0 & 1.2 & 0.6 & 1.0 & 3.0 \\
& GOT & 21.8 & 0 & 1.2 & 0.6 & 2.6 & 4.8 \\
& Social Network & 39.4 & 0 & 5.0 & 1.8 & 4.8 & 6.2 \\
& Politician & 40.6 & 0 & 2.2 & 2.6 & 5.2 & 7.0 \\
& Expert & 16.8 & 0 & 0.4 & 0.6 & 1.6 & 6.0 \\
\hline
\multirow{9}{*}{\begin{sideways}\method{few-shot}\end{sideways}}
& Overall & 42.7 & 10.3 & \underline{23.9} & \underline{10.2} & 13.3 & 26.0 \\
\cline{2-8}
& Adjacency & 50.2 & 11.8 & \textbf{77.6} & \textbf{27.0} & 17.4 & 83.4 \\
& Incident & 46.6 & 12.8 & 58.4 & 19.8 & 18.4 & 57.8 \\
& Co-authorship & 44.2 & 7.6 & 31.0 & 11.8 & 11.4 & 31.0 \\
& Friendship & 42.8 & 9.6 & 8.8 & 7.4 & 11.8 & 7.2 \\
& SP & 29.4 & 10.4 & 9.6 & 4.6 & 11.6 & 7.0 \\
& GOT & 26.0 & 10.0 & 8.2 & 5.4 & 9.0 & 9.8 \\
& Social Network & 40.4 & 9.4 & 8.4 & 4.2 & 12.0 & 11.2 \\
& Politician & 50.6 & 8.2 & 7.2 & 6.0 & 12.6 & 12.0 \\
& Expert & 54.0 & 12.6 & 6.0 & 6.0 & 15.8 & 14.6 \\
\hline
\multirow{9}{*}{\begin{sideways}\method{cot}\end{sideways}}
& Overall & \underline{50.6} & \underline{24.7} & 22.8 & 9.3 & 13.3 & \underline{77.0} \\
\cline{2-8}
& Adjacency & 51.0 & \textbf{80.8} & 72.2 & 22.0 & 19.6 & \textbf{84.0} \\
& Incident & 48.6 & 55.0 & 54.4 & 17.2 & 17.2 & 81.6 \\
& Co-authorship & 51.4 & 31.2 & 29.8 & 10.0 & 12.2 & 80.6 \\
& Friendship & 50.4 & 8.8 & 8.4 & 4.2 & 10.2 & 77.4 \\
& SP & 52.2 & 9.4 & 10.0 & 6.6 & 12.0 & 74.6 \\
& GOT & 51.4 & 9.2 & 8.0 & 5.6 & 10.4 & 70.4 \\
& Social Network & \textbf{53.8} & 8.6 & 7.8 & 5.8 & 8.4 & 76.0 \\
& Politician & 47.0 & 9.4 & 8.4 & 6.4 & 13.8 & 75.8 \\
& Expert & 49.6 & 10.2 & 6.4 & 5.6 & 15.6 & 72.6 \\
\hline
\multirow{9}{*}{\begin{sideways}\method{cot-bag}\end{sideways}}
& Overall & 50.3 & 23.7 & 19.4 & 9.2 & \underline{13.6} & 68.4 \\
\cline{2-8}
& Adjacency & 49.6 & 73.0 & 57.8 & 22.8 & \textbf{17.6} & 82.4 \\
& Incident & 49.6 & 53.4 & 46.6 & 17.6 & 14.4 & 77.2 \\
& Co-authorship & 50.4 & 30.4 & 28.4 & 10.2 & 14.6 & 74.0 \\
& Friendship & 48.8 & 8.4 & 6.2 & 5.2 & 9.2 & 65.8 \\
& SP & 50.4 & 7.0 & 6.8 & 5.0 & 12.4 & 61.2 \\
& GOT & 51.8 & 11.0 & 6.0 & 5.0 & 13.2 & 57.2 \\
& Social Network & 55.8 & 10.6 & 9.4 & 4.6 & 11.0 & 59.4 \\
& Politician & 49.2 & 9.4 & 6.0 & 6.6 & 16.0 & 69.6 \\
& Expert & 47.4 & 10.2 & 7.4 & 6.2 & 14.2 & 68.6 \\
\hline
\end{tabular}
}
\caption{Comparing different graph encoding functions on different graph tasks for PaLM 2 XXS. The most effective prompting heuristic is highlighted with an underline, and the top-performing graph function encoder for the respective heuristic is highlighted in bold.}
\label{table:graph-encoders-palm2xxs}
\end{table}

\begin{table}
\centering
\footnotesize
\setlength{\tabcolsep}{2pt}
\resizebox{\columnwidth}{!}{%
\setlength{\tabcolsep}{3pt}
\begin{tabular}{c|c|c|c|c|c|c|c}
Method & Encoding function & Edge Existence & Node degree & Node count & Edge count & Connected nodes  & Cycle check \\ \hline
\multirow{9}{*}{\begin{sideways}\method{zero-shot}\end{sideways}}
& Overall & \underline{56.6} & 11.1 & 11.7 & 3.9 & 8.2 &  20.8 \\
\cline{2-8}
& Adjacency & 57.6 & 10.4 & 10.0 & 4.6 & 15.6 & 23.4 \\
& Incident & 59.8 & 12.2 & 11.8 & 4.2 & 11.6 &  32.4 \\
& Co-authorship & 58.2 & 10.8 & 9.8 & 5.0 & 6.8 &  25.8 \\
& Friendship & 57.2 & 11.0 & 11.6 & 2.6 & 3.0 &  18.4 \\
& SP & 54.8 & 11.4 & 15.4 & 3.0 & 5.8 & 17.0 \\
& GOT & 51.2 & 9.0 & 16.2 & 3.4 & 5.2 & 16.6 \\
& Social Network & 54.4 & 11.8 & 11.0 & 3.2 & 6.6 &  15.2 \\
& Politician & 55.8 & 12.2 & 8.0 & 4.8 & 7.6 & 18.4 \\
& Expert & \textbf{60.2} & 11.0 & 11.8 & 4.0 & 11.4 & 19.6 \\
\hline
\multirow{9}{*}{\begin{sideways}\method{zero-cot}\end{sideways}}
& Overall & 45.9 & 17.6 & 11.2 & \underline{9.7} & 14.4 & 33.9 \\
\cline{2-8}
& Adjacency & 51.2 & 26.2 & 5.2 & \textbf{16.6} & 19.2 & 70.6 \\
& Incident & 52.4 & 38.8 & 6.0 & 13.6 & 16.8 & 46.4 \\
& Co-authorship & 47.0 & 25.0 & 10.8 & 12.4 & 13.8 & 32.8 \\
& Friendship & 47.2 & 10.0 & 16.6 & 9.2 & 11.4 &  21.4 \\
& SP & 39.4 & 9.0 & 12.8 & 9.2 & 14.2 & 22.6 \\
& GOT & 40.8 & 9.4 & 11.6 & 7.8 & 11.0 & 17.4 \\
& Social Network & 47.2 & 10.0 & 12.4 & 7.2 & 11.6 & 15.2 \\
& Politician & 48.2 & 13.0 & 13.6 & 6.2 & 12.8 &  22.8 \\
& Expert & 39.8 & 16.8 & 11.6 & 4.8 & 19.0 & 56.2 \\
\hline
\multirow{9}{*}{\begin{sideways}\method{few-shot}\end{sideways}}
& Overall & 54.1 & 10.0 & 11.9 & 4.9 & 8.6 & 82.6 \\
\cline{2-8}
& Adjacency & 53.2 & 11.2 & 16.2 & 4.6 & 9.2 & 82.8 \\
& Incident & 53.2 & 11.4 & 23.0 & 5.6 & 9.0 & 80.2 \\
& Co-authorship & 54.2 & 10.8 & 13.0 & 4.6 & 8.4 & 84.4 \\
& Friendship & 56.0 & 8.4 & 9.4 & 4.8 & 8.4 & 81.0 \\
& SP & 59.0 & 9.0 & 10.4 & 5.4 & 8.8 & 84.2 \\
& GOT & 52.8 & 8.8 & 9.4 & 5.8 & 8.6 & 84.6 \\
& Social Network & 50.6 & 9.8 & 8.8 & 4.2 & 7.8 & 85.4 \\
& Politician & 51.4 & 7.6 & 7.8 & 5.2 & 9.4 & 80.2 \\
& Expert & 56.6 & 12.8 & 9.2 & 3.8 & 8.0 & 80.8 \\
\hline
\multirow{9}{*}{\begin{sideways}\method{cot}\end{sideways}}
& Overall & 56.4 & \underline{20.0} & 17.3 & 6.6 & 8.0 & \underline{82.7} \\
\cline{2-8}
& Adjacency & 57.4 & \textbf{51.6} & 31.8 & 14.8 & 10.4 & 80.0 \\
& Incident & 56.4 & 41.0 & 37.0 & 10.8 & 11.2 & 80.8 \\
& Co-authorship & 54.4 & 28.6 & 19.6 & 7.2 & 7.6 & 83.4 \\
& Friendship & 58.0 & 11.0 & 11.6 & 4.0 & 6.0 & 81.6 \\
& SP & 62.4 & 10.4 & 12.2 & 3.4 & 5.4 & 84.4 \\
& GOT & 54.4 & 11.2 & 13.2 & 4.2 & 5.4 & 84.0 \\
& Social Network & 56.0 & 8.2 & 12.4 & 3.8 & 6.0 & \textbf{86.0} \\
& Politician & 53.8 & 8.4 & 9.6 & 6.4 & 8.2 & 80.8 \\
& Expert & 55.2 & 9.8 & 8.6 & 5.0 & 11.6 & 83.0 \\
\hline
\multirow{9}{*}{\begin{sideways}\method{cot-bag}\end{sideways}}
& Overall & 52.6 & 19.6 & \underline{17.7} & 8.1 & 8.1 & 82.1 \\
\cline{2-8}
& Adjacency & 53.4 & 50.2 & 30.4 & 17.4 & 10.2 & 81.2 \\
& Incident & 49.4 & 44.0 & \textbf{33.2} & 16.6 & 9.0 & 79.8 \\
& Co-authorship & 52.4 & 26.4 & 22.6 & 8.4 & 7.0 & 85.2 \\
& Friendship & 50.8 & 10.2 & 12.6 & 4.6 & 5.2 & 80.0 \\
& SP & 56.6 & 8.8 & 13.4 & 4.0 & 5.8 & 82.4 \\
& GOT & 52.6 & 11.0 & 12.8 & 5.4 & 6.4 & 82.4 \\
& Social Network & 54.0 & 8.0 & 12.6 & 5.4 & 7.8 & 82.2 \\
& Politician & 51.0 & 9.2 & 10.8 & 5.0 & 9.2 & 83.0 \\
& Expert & 53.2 & 9.0 & 10.6 & 5.8 & 12.6 & 82.8 \\
\hline
\end{tabular}
}
\caption{Comparing different graph encoding functions on different graph tasks for Flan-PaLM 2 XXS. The most effective prompting heuristic is highlighted with an underline, and the top-performing graph function encoder for the respective heuristic is highlighted in bold.}
\label{table:graph-encoders-palm2xxsf}
\end{table}




\begin{table}
\centering
\footnotesize
\setlength{\tabcolsep}{2pt}
\resizebox{\columnwidth}{!}{%
\setlength{\tabcolsep}{3pt}
\begin{tabular}{c|c|c|c|c|c|c|c}
Method & Encoding function & Edge Existence & Node degree & Node count & Edge count & Connected nodes & Cycle check \\ \hline
\multirow{9}{*}{\begin{sideways}\method{zero-shot}\end{sideways}}
& Overall & 49.9 & 23.0 & 28.7 & \underline{21.3} & 10.1 & 15.6 \\
\cline{2-8}
& Adjacency & 50.8 & 22.4 & 11.4 & 22.2 & 25.8 &  21.8 \\
& Incident & 50.6 & 36.6 & 11.2 & 11.0 & 31.0 & 30.6 \\
& Co-authorship & 48.2 & 21.4 & 31.0 & 17.8 & 11.6 & 6.8 \\
& Friendship & 49.2 & 20.6 & 36.2 & 24.0 & 4.2 & 0 \\
& SP & 52.4 & 22.6 & 38.4 & \textbf{25.8} & 2.8 &  0.4 \\
& GOT & 53.8 & 17.8 & 32.2 & \textbf{25.8} & 1.8 &  0 \\
& Social Network & 44.6 & 22.4 & 35.8 & 24.4 & 2.0 & 0 \\
& Politician & 49.0 & 21.4 & 32.8 & 21.8 & 5.2 & 15.2 \\
& Expert & 50.2 & 22.0 & 29.6 & 18.6 & 6.6 & 65.4 \\
\hline
\multirow{9}{*}{\begin{sideways}\method{zero-cot}\end{sideways}}
& Overall & 43.7 & 3.8 & 16.9 & 9.9 & 15.2 & 14.2 \\
\cline{2-8}
& Adjacency & 48.2 & 16.6 & 3.0 & 18.4 & 26.6 & 40.4 \\
& Incident & 53.6 & 10.6 & 2.6 & 6.2 & 50.8 & 43.6 \\
& Co-authorship & 46.8 & 2.6 & 9.2 & 2.2 & 19.6 & 8.6 \\
& Friendship & 32.8 & 0.4 & 18.4 & 12.6 & 3.2 & 3.6 \\
& SP & 40.2 & 0.4 & 21.4 & 4.0 & 4.8 & 7.6 \\
& GOT & 41.2 & 0.2 & 20.8 & 3.8 & 3.6 & 2.8 \\
& Social Network & 47.2 & 0 & 25.4 & 12.4 & 3.8 & 1.4 \\
& Politician & 47.0 & 0.8 & 27.6 & 15.6 & 4.8 & 10.0 \\
& Expert & 36.0 & 3.0 & 23.8 & 14.0 & 19.4 & 10.2 \\
\hline
\multirow{9}{*}{\begin{sideways}\method{few-shot}\end{sideways}}
& Overall & 40.4 & 22.3 & 26.0 & 18.7 & 16.5 & 27.2 \\
\cline{2-8}
& Adjacency & 42.2 & 23.2 & 43.2 & 29.6 & 21.4 & 58.0 \\
& Incident & 48.6 & 35.4 & 58.8 & 31.8 & 34.0 & 41.2 \\
& Co-authorship & 42.6 & 24.0 & 22.2 & 18.2 & 15.2 & 29.4 \\
& Friendship & 45.0 & 17.4 & 16.8 & 13.8 & 13.2 & 18.2 \\
& SP & 36.6 & 23.6 & 16.2 & 16.0 & 13.0 & 20.2 \\
& GOT & 32.4 & 19.6 & 17.2 & 17.8 & 10.6 & 17.0 \\
& Social Network & 41.6 & 20.8 & 18.4 & 14.4 & 12.8 & 21.0 \\
& Politician & 43.0 & 16.2 & 17.6 & 12.8 & 11.6 & 26.0 \\
& Expert & 31.4 & 20.2 & 23.2 & 13.8 & 17.0 & 13.4 \\
\hline
\multirow{9}{*}{\begin{sideways}\method{cot}\end{sideways}}
& Overall & 57.8 & \underline{30.2} & 28.2 & 17.0 & 19.7 & 36.4 \\
\cline{2-8}
& Adjacency & 43.4 & 63.0 & 43.0 & 26.8 & 33.0 & 69.8 \\
& Incident & 55.8 & \textbf{63.8} & 54.4 & 24.6 & 44.2 & 38.8 \\
& Co-authorship & 59.6 & 27.2 & 25.2 & 13.4 & 17.2 & 40.6 \\
& Friendship & 64.2 & 19.0 & 20.2 & 12.8 & 13.0 & 40.8 \\
& SP & 62.0 & 19.2 & 18.0 & 16.6 & 15.0 & 10.2 \\
& GOT & 62.4 & 19.6 & 20.6 & 17.4 & 12.2 & 6.2 \\
& Social Network & 61.0 & 21.4 & 23.0 & 13.2 & 10.4 & 42.6 \\
& Politician & 55.2 & 18.4 & 21.4 & 14.0 & 13.6 & 61.4 \\
& Expert & 56.4 & 20.0 & 27.6 & 14.4 & 18.8 & 17.4 \\
\hline
\multirow{9}{*}{\begin{sideways}\method{cot-bag}\end{sideways}}
& Overall & \underline{58.9} & 29.6 & \underline{30.0} & 15.8 & \underline{20.0} & \underline{37.1} \\
\cline{2-8}
& Adjacency & 49.8 & 57.8 & 43.0 & 26.4 & 32.8 & \textbf{71.2} \\
& Incident & 57.4 & 61.8 & \textbf{50.0} & 23.8 & \textbf{41.8} & 55.8 \\
& Co-authorship & 59.0 & 27.0 & 25.8 & 15.6 & 17.6 & 34.8 \\
& Friendship & \textbf{66.2} & 22.6 & 22.6 & 10.0 & 10.2 & 38.2 \\
& SP & 61.2 & 18.4 & 23.8 & 15.2 & 13.4 & 15.6 \\
& GOT & 61.2 & 20.4 & 27.2 & 15.0 & 13.6 & 9.8 \\
& Social Network & 60.6 & 19.2 & 24.8 & 10.8 & 13.4 & 35.8 \\
& Politician & 54.8 & 17.8 & 23.2 & 11.2 & 15.4 & 53.6 \\
& Expert & 60.0 & 21.4 & 29.6 & 14.4 & 21.8 & 19.2 \\
\hline
\end{tabular}
}
\caption{Comparing different graph encoding functions on different graph tasks for PaLM 2 XS. The most effective prompting heuristic is highlighted with an underline, and the top-performing graph function encoder for the respective heuristic is highlighted in bold.}
\label{table:graph-encoders-palm2xs}
\end{table}

\begin{table}
\centering
\footnotesize
\setlength{\tabcolsep}{2pt}
\resizebox{\columnwidth}{!}{%
\setlength{\tabcolsep}{3pt}
\begin{tabular}{c|c|c|c|c|c|c|c}
Method & Encoding function & Edge Existence & Node degree & Node count & Edge count & Connected nodes  & Cycle check \\ \hline
\multirow{9}{*}{\begin{sideways}\method{zero-shot}\end{sideways}}
& Overall & 68.4 & 10.2 & 26.8 & 4.4 & 23.0 & 84.4 \\
\cline{2-8}
& Adjacency & 78.0 & 17.6 & 39.0 & 7.2 & 34.4 & 87.2 \\
& Incident & 76.2 & 29.6 & 46.0 & 3.8 & 45.8 & 84.4 \\
& Co-authorship & 64.8 & 11.2 & 23.6 & 3.4 & 20.8 & 84.6 \\
& Friendship & 63.4 & 5.4 & 23.4 & 3.2 & 15.2 & 84.0 \\
& SP & 59.2 & 5.4 & 16.8 & 2.8 & 16.0 & 84.0 \\
& GOT & 62.6 & 4.6 & 19.6 & 3.2 & 18.2 & 83.2 \\
& Social Network & 72.0 & 4.4 & 17.6 & 4.0 & 17.8 & 84.0 \\
& Politician & 69.0 & 5.6 & 20.0 & 4.4 & 17.2 & 84.6 \\
& Expert & 70.6 & 7.6 & 35.4 & 7.2 & 21.4 & 84.0 \\
\hline
\multirow{9}{*}{\begin{sideways}\method{zero-cot}\end{sideways}}
& Overall & 54.3 & 16.4 & \underline{32.2} & 13.4 & \underline{25.1} & 59.9 \\
\cline{2-8}
& Adjacency & 68.6 & 34.8 & 23.0 & 16.6 & 36.8 & 82.0 \\
& Incident & 59.4 & 51.2 & 24.2 & 11.0 & \textbf{55.8} & 67.4 \\
& Co-authorship & 51.8 & 15.0 & 25.8 & 11.8 & 26.6 & 41.4 \\
& Friendship & 53.8 & 6.2 & 41.6 & 12.2 & 19.6 & 70.6 \\
& SP & 49.4 & 5.8 & 33.6 & 13.8 & 14.4 & 37.0 \\
& GOT & 47.0 & 7.6 & 27.6 & 12.8 & 16.6 & 38.6 \\
& Social Network & 51.4 & 8.0 & 34.4 & 12.4 & 18.4 & 74.0 \\
& Politician & 55.6 & 8.8 & 35.6 & 12.8 & 14.6 & 53.8 \\
& Expert & 51.8 & 10.6 & \textbf{44.2} & 17.0 & 23.2 & 74.6 \\
\hline
\multirow{9}{*}{\begin{sideways}\method{few-shot}\end{sideways}}
& Overall & 70.9 & 13.2 & 21.4 & 10.0 & 10.4 & 87.2 \\
\cline{2-8}
& Adjacency & 72.0 & 22.4 & 33.2 & 14.4 & 12.2 & 86.6 \\
& Incident & 81.8 & 27.0 & 33.6 & 7.2 & 22.0 & 83.4 \\
& Co-authorship & 68.0 & 17.8 & 20.2 & 11.2 & 8.0 & 89.4 \\
& Friendship & 66.8 & 7.2 & 17.0 & 9.0 & 4.8 & 86.8 \\
& SP & 67.6 & 7.0 & 15.8 & 10.0 & 6.0 & 88.6 \\
& GOT & 67.6 & 5.0 & 15.2 & 9.2 & 6.8 & 88.0 \\
& Social Network & 70.6 & 10.2 & 14.6 & 9.0 & 6.6 & 88.2 \\
& Politician & 71.2 & 8.6 & 16.4 & 9.0 & 6.6 & 87.6 \\
& Expert & 72.4 & 13.6 & 26.2 & 11.2 & 20.8 & 86.4 \\
\hline
\multirow{9}{*}{\begin{sideways}\method{cot}\end{sideways}}
& Overall & 71.9 & 23.8 & 20.7 & 12.6 & 14.0 & 86.7 \\
\cline{2-8}
& Adjacency & 76.0 & 72.4 & 30.2 & 25.6 & 16.6 & 85.4 \\
& Incident & 77.0 & 63.4 & 32.8 & 18.0 & 23.2 & 82.2 \\
& Co-authorship & 67.4 & 22.6 & 19.0 & 13.8 & 9.4 & 84.2 \\
& Friendship & 69.0 & 7.4 & 17.6 & 7.8 & 10.2 & 88.8 \\
& SP & 71.2 & 7.4 & 15.4 & 9.6 & 9.4 & 87.4 \\
& GOT & 71.0 & 9.8 & 16.4 & 7.4 & 10.2 & 89.4 \\
& Social Network & 75.0 & 6.8 & 11.4 & 8.0 & 8.0 & 88.4 \\
& Politician & 72.4 & 10.2 & 15.4 & 11.0 & 13.4 & 89.8 \\
& Expert & 68.2 & 14.6 & 28.4 & 11.8 & 25.2 & 84.8 \\
\hline
\multirow{9}{*}{\begin{sideways}\method{cot-bag}\end{sideways}}
& Overall & \underline{74.7} & \underline{25.0} & 27.9 & \underline{14.1} & 14.8 &  \underline{88.8} \\
\cline{2-8}
& Adjacency & 73.0 & \textbf{73.8} & 37.8 & \textbf{25.8} & 16.2 & 86.4 \\
& Incident & \textbf{80.0} & 63.4 & 35.0 & 19.8 & 23.4 & 84.2 \\
& Co-authorship & 72.4 & 25.2 & 28.8 & 13.8 & 11.2 & 86.0 \\
& Friendship & 74.6 & 8.6 & 25.0 & 11.4 & 8.6 & 90.4 \\
& SP & 74.4 & 9.0 & 23.6 & 10.6 & 9.6 & 90.0 \\
& GOT & 75.6 & 8.0 & 24.8 & 10.2 & 12.4 & \textbf{91.8} \\
& Social Network & 76.8 & 9.2 & 19.6 & 9.6 & 12.2 & 90.4 \\
& Politician & 71.2 & 12.6 & 22.2 & 11.8 & 13.4 & \textbf{91.8} \\
& Expert & 74.4 & 15.4 & 34.6 & 13.8 & 26.6 & 88.0 \\
\hline
\end{tabular}
}
\caption{Comparing different graph encoding functions on different graph tasks for Flan-PaLM 2 XS. The most effective prompting heuristic is highlighted with an underline, and the top-performing graph function encoder for the respective heuristic is highlighted in bold.}
\label{table:graph-encoders-palm2xsf}
\end{table}

\begin{table}
\centering
\footnotesize
\setlength{\tabcolsep}{2pt}
\resizebox{\columnwidth}{!}{%
\setlength{\tabcolsep}{3pt}
\begin{tabular}{c|c|c|c|c|c|c|c}
Method & Encoding function & Edge Existence & Node degree & Node count & Edge count & Connected nodes & Cycle check \\ \hline
\multirow{9}{*}{\begin{sideways}\method{zero-shot}\end{sideways}}
& Overall & 48.2 & 41.4 & 36.5 & 12.1 & \underline{25.0} &\underline{74.7} \\
\cline{2-8}
& Adjacency & 47.2 & 33.6 & 14.0 & 19.0 & 32.8 & \textbf{83.6} \\
& Incident & 44.0 & 68.8 & 15.0 & 19.8 & \textbf{72.2} & 82.2 \\
& Co-authorship & 49.2 & 36.2 & 41.6 & 12.6 & 14.0 & 57.8 \\
& Friendship & 50.4 & 36.8 & 46.6 & 8.6 & 17.0 & 83.4 \\
& SP & 50.0 & 35.6 & 51.6 & 8.0 & 18.2 & 40.6 \\
& GOT & 51.6 & 36.4 & 49.0 & 11.2 & 15.8 & 75.0 \\
& Social Network & 47.0 & 42.0 & 48.4 & 9.6 & 19.0 & 83.2 \\
& Politician & 49.0 & 41.2 & 37.8 & 7.6 & 14.2 & \textbf{83.6} \\
& Expert & 45.8 & 41.8 & 24.8 & 12.4 & 22.0 & 82.6 \\
\hline
\multirow{9}{*}{\begin{sideways}\method{zero-cot}\end{sideways}}
& Overall & 37.0 & 7.4 & 31.3 & 13.1 & 14.7 & 16.5 \\
\cline{2-8}
& Adjacency & 46.0 & 36.0 & 20.4 & 20.4 & 15.6 & 69.6 \\
& Incident & 51.2 & 26.8 & 7.8 & 12.0 & 72.4 & 46.6 \\
& Co-authorship & 53.6 & 0.6 & 44.2 & 9.8 & 16.2 & 5.8 \\
& Friendship & 34.6 & 0.4 & 36.4 & 15.8 & 5.2 & 4.8 \\
& SP & 43.8 & 2.6 & 44.6 & 14.8 & 5.6 & 0.6 \\
& GOT & 18.0 & 0 & 43.0 & 12.4 & 2.8 & 0.2 \\
& Social Network & 35.2 & 0 & 38.0 & 22.6 & 2.6 & 3.4 \\
& Politician & 30.4 & 0 & 30.2 & 7.0 & 2.2 & 1.4 \\
& Expert & 20.0 & 0.4 & 17.0 & 3.0 & 9.6 & 16.4 \\
\hline
\multirow{9}{*}{\begin{sideways}\method{few-shot}\end{sideways}}
& Overall & 47.5 & 36.9 & 40.0 & 16.9 & 22.9 & 40.4 \\
\cline{2-8}
& Adjacency & 61.4 & 37.2 & 81.2 & 18.4 & 37.8 & 43.8 \\
& Incident & 69.4 & 61.2 & 83.2 & 20.4 & 80.4 & 45.8 \\
& Co-authorship & 28.2 & 33.4 & 31.2 & 15.8 & 12.2 & 20.2 \\
& Friendship & 47.2 & 36.0 & 25.6 & 17.4 & 12.8 & 47.2 \\
& SP & 45.6 & 30.6 & 29.4 & 15.8 & 13.8 & 46.4 \\
& GOT & 26.2 & 29.2 & 27.4 & 18.2 & 12.0 & 37.8 \\
& Social Network & 50.4 & 35.8 & 30.8 & 16.6 & 13.8 & 42.4 \\
& Politician & 39.4 & 30.4 & 27.6 & 15.6 & 13.6 & 35.4 \\
& Expert & 60.0 & 38.2 & 23.4 & 14.2 & 10.0 & 45.0 \\
\hline
\multirow{9}{*}{\begin{sideways}\method{cot}\end{sideways}}
& Overall & \underline{57.6} & \underline{44.3} & 41.7 & 19.4 & 23.0 & 42.5 \\
\cline{2-8}
& Adjacency & 62.6 & 69.4 & 82.0 & 23.8 & 41.8 & 41.0 \\
& Incident & \textbf{68.2} & \textbf{78.4} & 80.8 & 26.6 & 79.6 & 44.4 \\
& Co-authorship & 46.8 & 36.0 & 35.2 & 16.6 & 12.8 & 34.6 \\
& Friendship & 66.2 & 38.0 & 29.0 & 17.4 & 9.6 & 46.4 \\
& SP & 66.4 & 32.4 & 29.2 & 17.6 & 13.8 & 45.6 \\
& GOT & 50.6 & 30.4 & 28.0 & 19.2 & 8.2 & 24.2 \\
& Social Network & 52.2 & 40.4 & 31.4 & 17.0 & 11.0 & 50.4 \\
& Politician & 43.6 & 32.2 & 30.8 & 16.4 & 8.0 & 44.8 \\
& Expert & 62.0 & 41.8 & 28.8 & 20.2 & 22.0 & 51.0 \\
\hline
\multirow{9}{*}{\begin{sideways}\method{cot-bag}\end{sideways}}
& Overall & 55.2 & 43.7 & \underline{44.4} & \underline{20.8} & 22.7 & 39.6 \\
\cline{2-8}
& Adjacency & 54.0 & 66.2 & 85.2 & \textbf{25.8} & 40.6 & 41.6 \\
& Incident & 59.4 & 77.2 & \textbf{89.0} & 24.8 & 81.6 & 42.8 \\
& Co-authorship & 41.8 & 38.4 & 35.0 & 18.8 & 11.8 & 29.4 \\
& Friendship & 66.0 & 36.8 & 31.8 & 22.2 & 6.6 & 48.8 \\
& SP & 63.8 & 30.2 & 31.0 & 17.8 & 10.2 & 33.8 \\
& GOT & 59.2 & 32.2 & 31.0 & 19.4 & 10.0 & 21.4 \\
& Social Network & 49.8 & 39.4 & 34.0 & 20.6 & 12.2 & 47.6 \\
& Politician & 47.8 & 32.4 & 32.8 & 16.8 & 6.2 & 38.6 \\
& Expert & 55.4 & 40.6 & 30.0 & 21.0 & 25.4 & 52.8 \\
\hline
\end{tabular}
}
\caption{Comparing different graph encoding functions on different graph tasks for PaLM 2 S. The most effective prompting heuristic is highlighted with an underline, and the top-performing graph function encoder for the respective heuristic is highlighted in bold.}
\label{table:graph-encoders-palm2s}
\end{table}

\begin{table}
\centering
\footnotesize
\setlength{\tabcolsep}{2pt}
\resizebox{\columnwidth}{!}{%
\setlength{\tabcolsep}{3pt}
\begin{tabular}{c|c|c|c|c|c|c|c}
Method & Encoding function & Edge Existence & Node degree & Node count & Edge count & Connected nodes & Cycle check \\ \hline
\multirow{9}{*}{\begin{sideways}\method{zero-shot}\end{sideways}}
& Overall & 47.5 & 55.1 & \underline{76.3} & 30.6 & 19.5 & \underline{83.3} \\
\cline{2-8}
& Adjacency & 43.6 & 49.6 & \textbf{100.0} & 36.8 & 55.6 &\textbf{83.8} \\
& Incident & 48.6 & 85.0 & 98.6 & 6.6 & 88.0 & 83.2 \\
& Co-authorship & 48.0 & 55.2 & 67.4 & 32.4 & 1.6 & 83.2 \\
& Friendship & 48.0 & 50.8 & 63.2 & 31.2 & 0.2 & 83.2 \\
& SP & 49.2 & 49.8 & 56.4 & 30.8 & 6.6 & 83.2 \\
& GOT & 51.0 & 52.6 & 70.6 & 34.8 & 6.2 & 83.2 \\
& Social Network & 46.0 & 50.2 & 61.0 & 31.6 & 0 & 83.2 \\
& Politician & 50.0 & 52.8 & 70.8 & 32.2 & 1.0 & 83.2 \\
& Expert & 43.0 & 50.2 & 98.6 & 39.4 & 16.0 & 83.2 \\
\hline
\multirow{9}{*}{\begin{sideways}\method{zero-cot}\end{sideways}}
& Overall & 41.6 & 7.9 & 73.9 & 24.4 & 39.5 & 22.4 \\
\cline{2-8}
& Adjacency & 31.6 & 13.2 & 66.4 & 25.2 & 61.6 & 52.0 \\
& Incident & 52.0 & 54.8 & 75.0 & 9.8 & 84.4 & 55.2 \\
& Co-authorship & 43.4 & 0.8 & 81.8 & 31.6 & 27.8 & 9.0 \\
& Friendship & 46.0 & 0.6 & 80.2 & 26.2 & 20.4 & 8.4 \\
& SP & 38.6 & 0 & 78.6 & 27.6 & 41.0 & 0.2 \\
& GOT & 38.8 & 0.4 & 68.6 & 30.0 & 29.8 & 1.0 \\
& Social Network & 47.8 & 0 & 78.4 & 30.2 & 28.2 & 6.4 \\
& Politician & 51.4 & 1.2 & 66.6 & 29.2 & 17.2 & 11.6 \\
& Expert & 24.8 & 0.2 & 69.4 & 9.6 & 45.0 & 57.4 \\
\hline
\multirow{9}{*}{\begin{sideways}\method{few-shot}\end{sideways}}
& Overall & 41.5 & 55.8 & 60.3 & \underline{35.9} & \underline{46.1} & 73.8 \\
\cline{2-8}
& Adjacency & 49.2 & 61.0 & 97.6 & \textbf{43.2} & 66.4 & 78.4 \\
& Incident & 73.2 & 82.4 & 99.2 & 37.4 & \textbf{85.6} & 78.4 \\
& Co-authorship & 16.4 & 51.4 & 45.4 & 32.2 & 36.8 & 73.6 \\
& Friendship & 32.6 & 53.0 & 45.6 & 35.2 & 44.2 & 79.4 \\
& SP & 33.2 & 45.6 & 40.6 & 35.2 & 30.6 & 57.8 \\
& GOT & 30.0 & 48.6 & 41.6 & 38.2 & 36.0 & 61.4 \\
& Social Network & 40.4 & 51.4 & 44.0 & 32.4 & 38.2 & 79.0 \\
& Politician & 39.4 & 53.6 & 46.2 & 35.2 & 32.2 & 77.6 \\
& Expert & 59.2 & 55.0 & 82.8 & 34.0 & 45.0 & 78.6 \\
\hline
\multirow{9}{*}{\begin{sideways}\method{cot}\end{sideways}}
& Overall & 52.2 & 59.7 & 62.2 & 34.4 & 45.2 & 72.7 \\
\cline{2-8}
& Adjacency & 53.6 & 81.4 & 98.0 & 42.2 & 66.6 & 66.8 \\
& Incident & 72.4 & 94.6 & 98.8 & 29.8 & 87.2 & 68.6 \\
& Co-authorship & 42.4 & 55.0 & 45.0 & 33.8 & 37.0 & 69.4 \\
& Friendship & 55.0 & 48.8 & 45.0 & 33.4 & 40.8 & 80.6 \\
& SP & 54.4 & 48.4 & 44.6 & 34.2 & 26.2 & 71.0 \\
& GOT & 51.4 & 51.2 & 46.2 & 35.2 & 29.4 & 65.4 \\
& Social Network & 42.8 & 51.4 & 43.6 & 32.4 & 39.8 & 77.0 \\
& Politician & 36.8 & 53.0 & 50.4 & 33.0 & 30.2 & 76.8 \\
& Expert & 60.8 & 53.2 & 88.4 & 35.8 & 49.6 & 78.6 \\
\hline
\multirow{9}{*}{\begin{sideways}\method{cot-bag}\end{sideways}}
& Overall & \underline{60.4} & \underline{60.0} & 63.1 & 34.6 & 45.0 & 69.2 \\
\cline{2-8}
& Adjacency & 64.6 & 78.0 & 98.6 & 39.4 & 64.2 & 70.4 \\
& Incident & \textbf{71.6} & \textbf{95.4} & 99.4 & 32.2 & 89.0 & 70.8 \\
& Co-authorship & 52.0 & 55.8 & 48.6 & 33.0 & 36.8 & 63.0 \\
& Friendship & 64.8 & 49.6 & 45.6 & 32.6 & 36.4 & 71.8 \\
& SP & 65.8 & 51.6 & 44.2 & 33.0 & 26.4 & 69.4 \\
& GOT & 65.0 & 50.2 & 44.6 & 38.6 & 29.6 & 64.0 \\
& Social Network & 48.4 & 53.6 & 46.4 & 31.2 & 45.6 & 72.0 \\
& Politician & 50.4 & 53.0 & 51.0 & 33.8 & 24.2 & 63.4 \\
& Expert & 61.2 & 52.6 & 89.6 & 37.2 & 52.8 & 78.4 \\
\hline
\end{tabular}
}
\caption{Comparing different graph encoding functions on different graph tasks for PaLM 2 L. The most effective prompting heuristic is highlighted with an underline, and the top-performing graph function encoder for the respective heuristic is highlighted in bold.}
\label{table:graph-encoders-palm2l}
\end{table}

\end{document}